\documentclass{bmvc2k}
\hypersetup{hypertexnames=false}
\usepackage{multirow}%
\usepackage{amssymb}%
\usepackage{todonotes}
\usepackage{booktabs}
\usepackage[table]{xcolor}
\usepackage{cleveref}

\usepackage{wrapfig}
\usepackage{lipsum}
\usepackage{amsmath}
\usepackage{graphicx}
\usepackage{float}
\usepackage{listings}

\lstdefinestyle{prolog}{
  language=Prolog,
  basicstyle=\scriptsize\ttfamily,
  numbers=left,
  numberstyle=\tiny\color[gray]{0.65},
  numbersep=4pt,
  xleftmargin=4em,
  backgroundcolor=\color{blue!2},
  breaklines=true,
  keepspaces=true,
  columns=fullflexible,
  frame=none,
}

\definecolor{headerblue}{RGB}{220, 232, 246}

\title{Visual Commonsense Driven Knowledge Refinements for Scene Graph Generation}

\addauthor{Maëlic Neau}{maelicneau@gmail.com}{1}
\addauthor{Salim Baloch}{s.salim@constructor.university}{2,4}
\addauthor{Jakob Suchan}{jsuchan@constructor.university}{2,4}
\addauthor{Zoe Falomir}{zoe.falomir@umu.se}{1}
\addauthor{Mehul Bhatt}{mehul.bhatt@oru.se}{3,4}

\addinstitution{
 Computing Science Department,\\
 Umeå University,\\
 Umeå, Sweden
}
\addinstitution{
 School of Computer Science \& Engineering,\\
 Constructor University,\\
 Bremen, Germany
}
\addinstitution{
 School of Science and Technology,\\
 Örebro University,\\
 Örebro, Sweden
}
\addinstitution{
 CoDesign Lab EU.,\\
 Örebro, Sweden
}

\runninghead{Neau et. al.}{Commonsense Driven Knowledge Refinements for SGG}

\def\eg{\emph{e.g}\bmvaOneDot}

\definecolor{headerblue}{RGB}{220, 232, 246}

\newcommand{\numItem}[1]{{{\color{black} \small\textbf{(#1)}}}}

\begin{document}

\maketitle

\begin{abstract}
Learning-driven \emph{Scene Graph Generation} (SGG) models excel on frequent relation types but degrade sharply under annotation sparsity, failing to capture reliable visual commonsense knowledge.  We propose a model-agnostic, semantically-guided knowledge refinement framework that systematically mines commonsense-grounded constraints from training data --capturing  \emph{spatial, functional, and qualitative relational regularities}-- and uses general declarative commonsense reasoning to correct and refine ranked SGG predictions at inference time. The framework requires no manual rule authoring, no model retraining, and transfers across datasets and architectures. On three standard benchmarks, we obtain consistent improvements over strong baselines, demonstrating that structured visual commonsense reasoning over deep scene semantics is a practical and effective complement to purely learning-based scene graph generation.

\end{abstract}

% \begin{abstract}

% Scene graphs are powerful representations that abstract the content of images and videos as graph structures, with nodes representing visual entities grounded to the scene and edges natural language predicates. The task of automatically extracting scene graphs from images is called Scene Graph Generation (SGG).
% This task is important to support downstream applications that rely on compositional reasoning such as VQA or Image Generation. In SGG, learning-based methods have been the dominant approaches, with solutions adapted based on popular vision framework such as Faster-RCNN or DETR. Even if efficient on leading benchmarks, fully learning-based approaches can fail to learn visual commonsense when data is scarce. In this work, we propose to use visual abduction in Answer-Set Programming (ASP) to correct and enlarge predictions of a given SGG model. By automatically mining commonsense-grounded constraints from large datasets, we present a robust refinement framework for Scene Graph Generation. Our approach can generalize across datasets and models, benefiting any SGG models on standard benchmarks.

% \end{abstract}

%-------------------------------------------------------------------------
\section{Introduction}
\label{sec:intro}

Scene graphs \cite{krishna2017visual} are powerful representations that abstract visual scenes as collections of $\langle\text{subject},\text{predicate},\text{object}\rangle$ triplets. These triplets form a directed acyclic graph where subjects and objects become nodes grounded to the visual content by 2D or 3D coordinates (i.e. bounding boxes) \cite{armeni20193d}. Edges are predicates representing for instance actions (e.g. \textit{drinking from}) or spatial relations (e.g. \textit{behind}), even though formal definitions do not restrict predicates to specific ontologies or taxonomies \cite{krishna2017visual,amodeo2022og}. 
Scene graphs have shown great potential as a foundational element for various subsequent tasks that hinge on compositional aspects, such as Image Captioning \citep{yang2019auto}, Visual Question Answering (VQA) \citep{lee2019visual} or image generation \cite{johnson2018image}. However, manually annotating scene graphs is a resource-intensive process, and thus applications rely extensively on Scene Graph Generation (SGG) \cite{xu2017scene} methods to automatically predict objects and relations from images or videos.

\smallskip

\textbf{Scene Graph Learning}.\quad Scene graph generation methods are predominantly learning-based, modelling relations as a distribution of predicate classes over the combination of all possible pairs per image \cite{zellers2018neural,xu2017scene,Neau_2025_BMVC}. However, the training signal for these approaches is scarce, since only a small subset of all possible relations are annotated in practice in leading datasets \cite{krishna2017visual,yang2022panoptic}. The $\langle\text{subject},\text{predicate},\text{object}\rangle$ space is then combinatorial, however its supervision does not scale accordingly, e.g. from $C$ object classes and $P$ predicates, $O(C^2P)$ admissible triplets might be obtained. But in reality, the training annotations cover only a few tens of thousands of distinct combinations, each with a handful of examples \cite{krishna2017visual,yang2022panoptic}. Thus, the commonsense rules that govern these triplets  are present in the data only as weak aggregate signals scattered across hundreds of thousands of pair classifications. Recovering them through per-pair gradient updates is asking a \textit{softmax} function to triangulate a heavy-tailed combinatorial structure from a small handful of pixel-level observations per cell, all while competing with a head-class prior that pushes every ambiguous example toward the generic answer \cite{tang2020unbiased}. It is not a matter of including more capacity or more data of the same kind since the rules are confounded with the long-tail noise at the level of individual pair gradients, and no amount of end-to-end training reorganises that signal into the structural constraint it actually represents. To solve this limitation, commonsense rules can be enforced post-hoc, as a \textit{consistency layer} adjacent to model predictions.

\smallskip

\begin{figure}
    \centering
    \includegraphics[width=\linewidth]{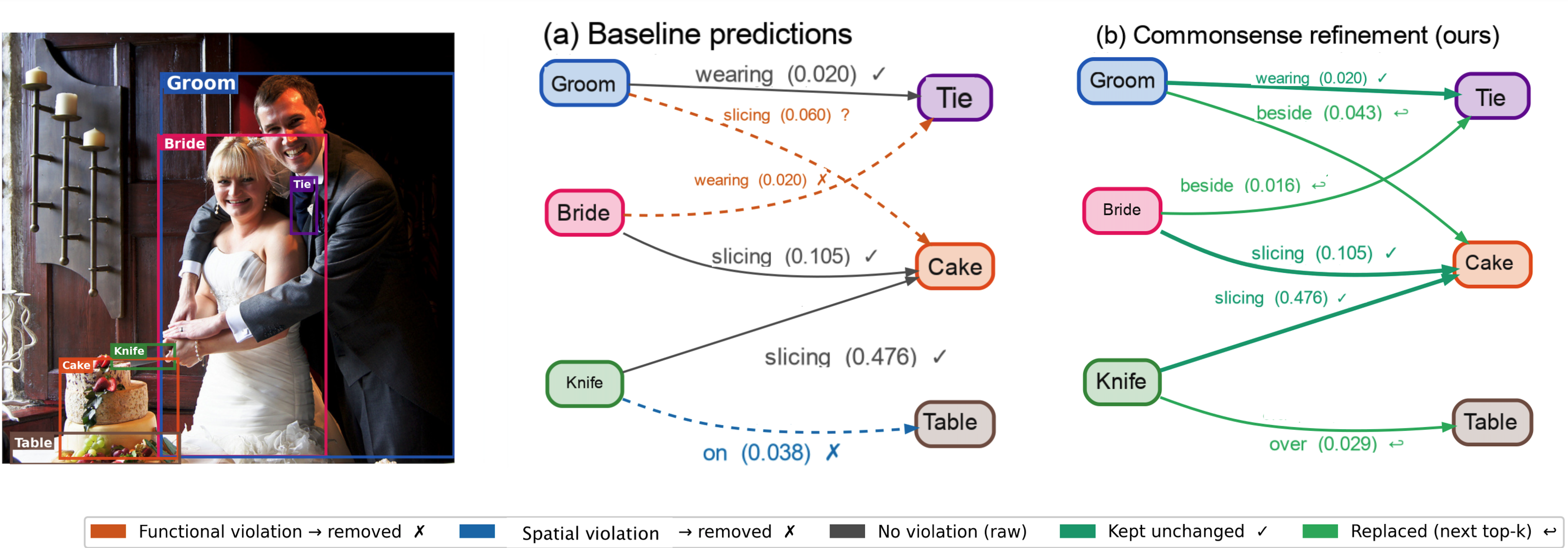}
    \caption{{\bfseries Sample Refinement Effects for a PSG image}. The baseline model (a) predicts three relations that violate commonsense constraints, either because of implausible functional (red) or spatial (blue) constraints. While our refinement (b) eliminates these violations and recovers two missing relations (green edges).}
    \label{fig:intro}
\end{figure}

\textbf{Commonsense Driven Scene Graph Refinment}.\quad Commonsense rules govern most of the space of visual relations \cite{davis2015commonsense,khan2025survey}. The relations in scene graph or predicates over domain entities are not arbitrary labels: they are linguistic proxies for \textit{spatial}, \textit{functional}, and \textit{qualitative relational} structure. A relation such as \emph{wearing} simultaneously encodes a containment geometry, a unitary ownership semantics, and relation-specific meta-property, e.g., an asymmetric directionality or a transitivity of physical containment. Learning-based SGG models collapse this multi-faceted structure into a flat per-pair score, conflating constraints that are qualitatively distinct in nature. We decompose visual commonsense into three orthogonal regularity classes, 
each reflecting a different mode by which the world constrains visual relations:

\textbullet \hspace{4pt}\textbf{Spatial Grounding}. Predicates are physically realised through the spatial configuration of their arguments --topology, containment, relative direction-- largely invariant to appearance. Violations are not statistical anomalies; they are physical impossibilities.

\textbullet \hspace{4pt}\textbf{Structural Cardinality}. Predicates typically encode roles that are capacity-bounded: a vehicle has \textit{one} driver, a garment \textit{one} wearer. Models that do not handle such constraints, e.g., when scoring pairs independently without an explicit mechanism to handle cardinalities, fail to adequately learn these role-cardinality regularities, e.g., freely admitting multiple simultaneous drivers for a single vehicle (see Fig. \ref{fig:qualitative_ex1})

%Such constraints are invisible to models that score pairs independently with no mechanism to reason over the global in-degree of their predictions.

\textbullet \hspace{4pt}\textbf{Logical Inference}. Predicates participate in inferential relationships independent of any image: symmetries, inverses, and compositions that follow from the semantics of the (relational) spatial calculus \cite{ligozat-qsr}. A model predicting \emph{above} without inferring \emph{below} is not merely inaccurate -- it is logically inconsistent.

These classes are orthogonal -- no one subsumes another-- and jointly exhaustive of the commonsense structure that is both recoverable from annotations and systematically enforceable through logical inference. Consider the example in Figure~\ref{fig:intro}. The baseline model \numItem{a} predicts three relations that violate commonsense constraints, either because of implausible functional (red) or spatial (blue) constraints. While our refinement \numItem{b} eliminates these violations and recovers two missing relations (green edges).

\smallskip

\textbf{Key Contributions}.\quad The key contributions of this paper are: \numItem{1} a unified, training-data-driven rule mining stage that extracts geometrically rooted spatial, functional, and relational commonsense priors requiring no manual annotation or external knowledge base;  \numItem{2} a systematic declarative commonsense reasoning method that operates post-hoc on SGG model output, filtering physically implausible predictions and recovering logically entailed but missing relations through formally verifiable constraint-based reasoning. \numItem{3}  a comprehensive empirical evaluation across three benchmarks (PSG, VG150, IndoorVG) and three SGG architectures (Motifs, Transformer, REACT++) demonstrating consistent F1@K gains, together with a quantification  of commonsense consistency with a Constraint Violation Rate (CVR) metric semantically capturing refinement benefits not captured by purely recall-based metrics. The method, while remaining model-agnostic and not requiring retraining, serves as a practical and transferable complement to learning-based SGG while also providing formally verifiable and explainable corrections for computed knowledge refinements.

\section{Related Work}\label{sec:related}
Our research integrates and advances methods from two research communities: \emph{scene graph generation} and visual commonsense as pursued within computer vision \cite{zellers2019recognition,khan2025survey}, and \emph{computational visual commonsense reasoning} as pursued within the knowledge representation and reasoning community within AI research \cite{Bhatt21_Artificial_Visual_Intelligence}:

\smallskip

\textbf{Scene Graph Generation}.\quad Scene graph generation methods model predicate distributions over pairwise  object combinations, treating commonsense regularities as emergent signals rather than explicit targets~\cite{xu2017scene,zellers2018neural,tang2020unbiased}. Prior work has approached this gap from the viewpoint of symbolic methods~\cite{khan2022expressive,amodeo2022og} enforcing structured constraints post-hoc via external knowledge graph infusion or manually authored ontology axioms demonstrating that prior knowledge meaningfully improves SGG, particularly in robustness-critical applications. The commonsense, however, is either externally sourced and misaligned with target dataset statistics, or hand-crafted and domain-specific, with no path to open-domain generalisation. \citet{zareian2020learning} instead learn visual commonsense directly from annotated data, akin to our own data-driven approach, but absorb it into model weights, leaving constraints unverifiable and logically inconsistent outputs uncorrected at inference time. Recently, \citet{jiang2025enhancing} post-process predictions using LLMs as commonsense oracles, the closest in spirit to our work, but without spatial or functional constraint enforcement, or abduction of missing relations, or systematic and general-purpose formal (declarative) verifiability.

\smallskip

\begin{figure}
    \centering
    \includegraphics[width=\linewidth]{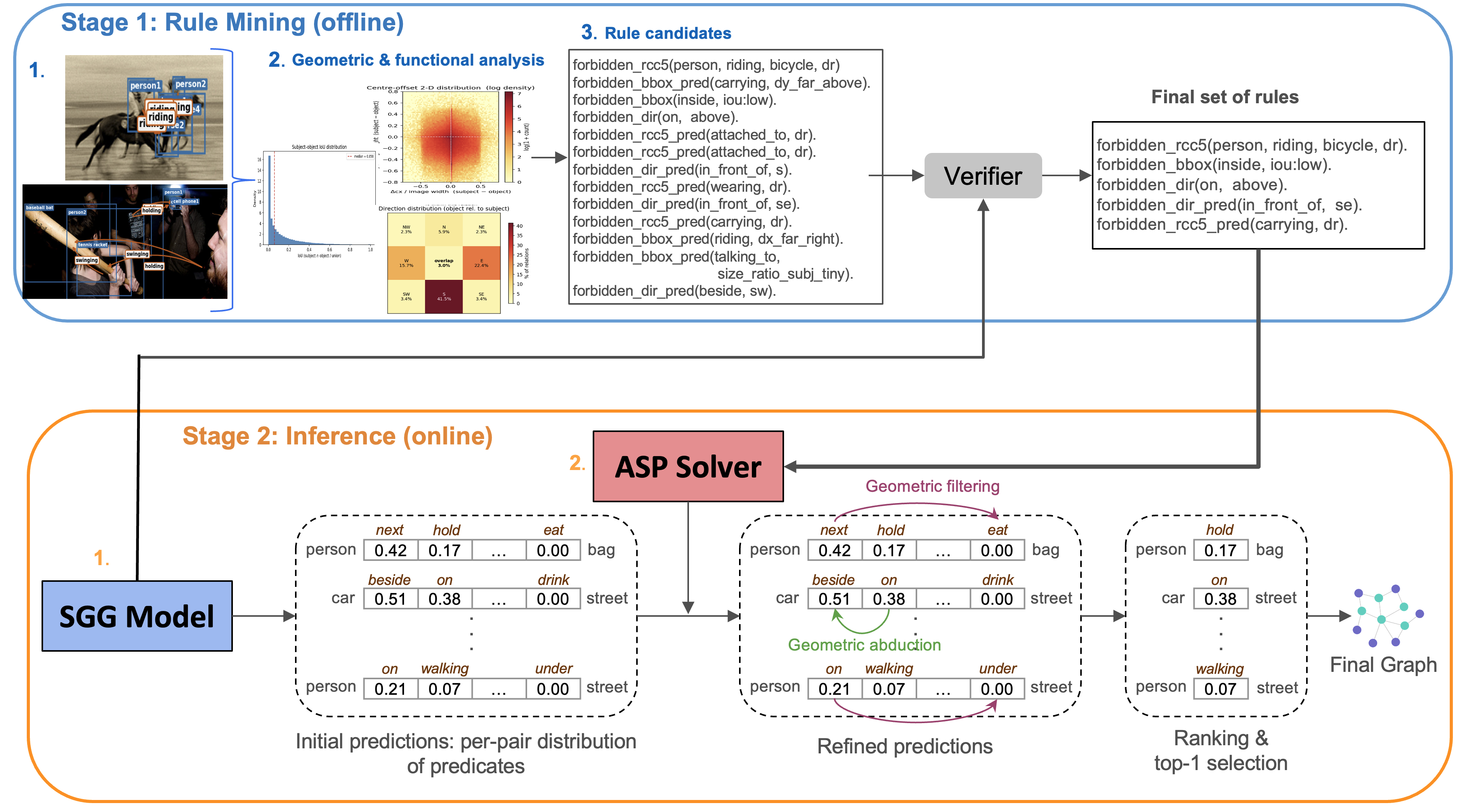}
    \caption{\textbf{Visual Commonsense Driven SGG}. Conceptual overview of the proposed knowledge refinement method.}
    \label{fig:approach}
\end{figure}

\textbf{Visual Commonsense Reasoning}.\quad Within research on computational commonsense knowledge representation and reasoning, the development of (neurosymbolic) visual commonsense methods have recently gained atraction as a promising line of research with the aim to achieve a general integration of state-of-the-art methods in knowledge representation and reasoning and (deep learning based) computer vision  \cite{sqa-suchan16,AIJ2021-Suchan,Yang2020_NeurASP,pmlr-v284-eiter25a,PADALKAR_GUPTA_2025} (see review in \cite{Bhatt21_Artificial_Visual_Intelligence}). In particular, the use of declarative techniques in logic and answer set programming (ASP) standout as sustained lines of work from viewpoints such as  (semantic) visual-question answering in logic and constraint logic settings \cite{sqa-suchan16,pmlr-v284-eiter25a,PADALKAR_GUPTA_2025}, and non-monotonic visual abduction for (neurosymbolic) spatio-temporal belief maintenance in dynamic domains within ASP settings \cite{AIJ2021-Suchan,Suchan2025_KR}. Differences in the technical framing and supported computational capabilities of these select exemplary works notwithstanding, the general motivation underlying remains unified: to  integrate powerful low-level visual processing capabilities to extract (geometric) scene elements and visual features from the imagery with high-level, expressive conceptual commonsense knowledge with the aim to support  neurosymbolic interpretation of either static and/or dynamic stimuli.

\smallskip

Some related works deal with injecting or infusing extra relations into the SGG \cite{KhanBC25,khan2022expressive} (e.g. affordances from embeddings or part-hole relations from ontologies) with the aim to enrich the SGG. Few works in the literature deal with the challenge to check the consistency of the SGG generated rules or its correctness. Jian et al. \cite{jiang2025enhancing} used LLMs to validate the obtained predicates (which sometimes hallucinate saying that \textit{finger of women} or \textit{window on bus} are physically impossible relations). Yuan et al. \cite{yuan2025commonsense} generated SGGs using LLMs (avoiding proper symbol grounding) for visual query answering. The intersection of structured (formal) commonsense reasoning and SGG remains sparsely covered~\cite{khan2025survey}.  Systematically addressing this integration, the proposed visual commonsense driven knowledge refinement methodology of this paper mines commonsense constraints directly from training data. We ensure that the mined constraints are explicit, and enforced through (formal) declarative reasoning without an external resource, retraining, or opaque oracle. The proposed method is general, declaratively formalised and robust to ensure a systematically enforceable treatment of the full commonsense constraint space guiding scene graph generation.

\section{Methodology}\label{sec:method}

Our approach is decoupled in two steps: first, we mine commonsense rules from evidence contained in the annotated data in an offline process and second, we use these rules to power declarative commonsense reasoning using Answer-Set Programming (ASP) at inference time to refined model's predictions. We summarize the overall architecture in Figure~\ref{fig:approach}.
During an offline mining stage, commonsense constraints and relational regularities are extracted from annotated scene graph datasets and compiled into a declarative ASP program. At inference time, the neural SGG predictions together with the spatial relations of candidate object pairs are translated into ASP facts. Abductive commonsense reasoning is then used to select a globally consistent refined scene graph.

\medskip

\textbf{Scene Graph Generation.} \quad
A \emph{scene graph} is a directed labelled graph $G = (\mathcal{O}, \mathcal{T})$ where $\mathcal{O}$ is a set of \emph{object instances} and $\mathcal{T} \subseteq \mathcal{O} \times \mathcal{P} \times \mathcal{O}$ is a set of \emph{relation triplets}, with $\mathcal{P}$ a fixed predicate vocabulary. Each instance $o \in \mathcal{O}$ carries a class label $\ell(o) \in \mathcal{C}$ and a bounding box $\beta(o) \in \mathbb{R}^4$. A Scene Graph Generation (SGG) model takes an image $I$ and produces, for each \texttt{<subject, object>} pair $(s, o) \in \mathcal{O} \times \mathcal{O}$ with $s \neq o$, a  \emph{ranked distribution over predicates}: a tuple $\sigma(s,o) = (p_1, \ldots, p_k)$ with associated scores $\mathbf{f}(s,o) = (f_1, \ldots, f_k)$ where $f_i \in [0,1]$ is the model's confidence that predicate $p_i$ holds between $s$ and $o$. At runtime, naive SGG typically selects the \emph{top-1} predicate per pair: $\hat{\mathcal{T}}_{\text{top-1}} = \{(s, p_1, o) : (s,o) \in \mathcal{O}^2, s \neq o\}$. Then, the final score per triplet is computed as a product of the predicate score and the object class scores: $f(s,p,o) = f_p(s,o) \cdot f_s(s) \cdot f_o(o)$, where $f_p(s,o)$ is the score of predicate $p$ for pair $(s,o)$, and $f_s(s)$ and $f_o(o)$ are the scores of the object detector for the subject $s$ and object $o$ bounding boxes, respectively. At runtime and during evaluation, the top-$k$ relations are selected per image, yielding to the final graph $\hat{G} = (\mathcal{O}, \hat{\mathcal{T}})$ where $\hat{\mathcal{T}} \subseteq \hat{\mathcal{T}}_{\text{top-1}}$ is the set of selected triplets.

%\subsection{Declarative Visual Commonsense}
%TODO

%Our approach is decoupled in two steps: first, we mine commonsense rules from evidence contained in the annotated data in an offline process and second, we use these rules to power declarative commonsense reasoning using Answer-Set Programming (ASP) at inference time to refined model's predictions. We summarize this architecture in Figure~\ref{fig:approach}.

\medskip

\textbf{Declarative Visual Commonsense.} \quad 
Visual commonsense reasoning  for scene graph refinement is based on declarative characterizations of relational rules and constraints in Answer Set Programming (ASP) \citep{Gebser2014-Clingo,ASP-books/sp/Lifschitz19,Gebser2016-clingo5}.\footnote{Answer Set Programming (ASP) is an established declarative programming paradigm for combinatorial search and constraint satisfaction within an expressively and computational robust non-monotonic (abductive) reasoning setting \cite{Gebser2014-Clingo,ASP-books/sp/Lifschitz19}.} 
%ASP is an established and widely applied foundational declarative language and robust computing methodology for a range of (non-monotonic) knowledge representation and reasoning tasks \cite{Gebser2012-ASP,Gebser2016-clingo5}. 
%  \cite{ASP-SI-2018,Gebser2012-ASP,Gebser2014-Clingo,ASP-Glance-2011,ASP-books/sp/Lifschitz19,Gebser2016-clingo5}
%Rooted in the \emph{stable model semantics} of logic programs, reasoning in ASP is equivalent to model-theoretic derivation and selection/optimisation of  interpretations, i.e., stable models or answer sets, under non-monotonic assumptions and support for defaults, exceptions, indirect effects, preferences. In essence, 
%
Most relevant for this paper is its support for declarative selection among alternative models via minimisation and optimisation, making it particularly suited to diverse forms of commonsense reasoning, e.g., abductive explanation in the presence of incomplete or noisy information \cite{AIJ2021-Suchan}. 
In ASP relational problems are modeled as answer set programs $\Pi$ which are finite sets of rules of the form: 
\qquad  \qquad $h \;\! \leftarrow \;\! b_1,\dots,b_m,\; \text{not } c_1,\dots,\text{not } c_n$, where $h$, $b_i$, and $c_j$ are (possibly negated) atoms. %, and $\text{not}$ denotes \emph{default negation}. 
Intuitively, the rule states that $h$ is derivable if all positive body literals $b_i$ hold and none of the negative literals $c_j$ can be derived. 
Solutions to such an ASP program $\Pi$ are given by its stable models or answer sets $\mathbf{SM}$[$\Pi$], which are self-supporting interpretations that satisfy all rules included in $\Pi$.
%
%\smallskip
%
%
%
%Let $\Pi$ be a ground ASP program and $I$ an interpretation. 
%
%%The {Gelfond--Lifschitz reduct} of $\Pi$ with respect to $I$ is obtained by:  {\small\textbf{(1)}} discarding every rule $r \in \Pi$ whose body contains a default-negated literal $\text{not } c$ with $c \in I$; and {\small\textbf{(2)}} removing all remaining default-negated literals from rule bodies. 
%
%%The resulting program $\Pi^I$ is a positive (negation-free) logic program. 
%
%An interpretation $I$ is a {stable model} or an {answer set}) of $\Pi$ if $I$ is a \emph{minimal model} (under set inclusion) of the reduct of $\Pi$ with respect to $I$. 
%We denote a stable model of an ASP program as $\mathbf{SM}$[$\Pi$]. Stable models constitute self-supporting interpretations that satisfy all rules under a non-monotonic reading. 
%
In addition to these standard rules, the following are of importance: %from the viewpoint of this paper:
(i) \textit{Choice Rules}, which allow the program to non-deterministically select among alternative hypotheses. In our setting, they are used to generate candidate relation assignments consistent with the confidence-ranked predictions of the neural SGG model.
%
%\smallskip
%
And (ii) \textit{Integrity Constraints}, which are rules with empty heads that eliminate invalid interpretations. They are used to enforce commonsense consistency conditions over candidate scene graphs.

\smallskip

%We model visual commonsense as a declarative background theory $\Sigma$ expressed as an ASP program composed of general rules describing geometric consistency conditions, functional cardinality constraints, and qualitative relational regularities, and specific background facts mined from annotated scene graph datasets. 
%During inference, $\Sigma$ is combined with the neural SGG predictions to guide abductive refinement under stable model semantics.

Visual commonsense is then modeled as a declarative ASP program composed of general rules describing spatial consistency conditions, functional cardinality constraints, and qualitative relational regularities, and specific background facts mined from annotated scene graph datasets. 
During inference, this is combined with the neural SGG predictions for abductive refinement based on the mined commonsense rules.

%We model visual commonsense as domain independent structural knowledge denoted by $\Sigma$, consisting of domain-independent structural constraints and inference rules compiled from statistical regularities mined from annotated scene graph datasets. Concretely, these include geometric constraints over spatial configurations, functional constraints over admissible relational cardinalities, and qualitative relational regularities capturing logical dependencies between predicates, implemented as a general set of rules captured in the ASP program $\Pi_{abd}$

%It is important to note that in this setting, reasoning is model-selective rather than proof-oriented, and that multiple stable models may exist, each corresponding to a distinct, coherent world model.

%\medskip

\subsection{Mining Commonsense Rules}
\label{sec:rule-mining}

%From the training annotations, we mine three independent families of rules that together summarise the spatial, structural, and logical regularities of the scene-graph predicates.

We mine three independent families of rules from the training annotations of a scene graph dataset $\mathcal{D}$, that together summarise the spatial (\textbf{I}.), functional (\textbf{II}.), and relational (\textbf{III}.) regularities of the scene-graph predicates and compile them into a dataset-specific ASP program $\Pi_{\mathcal{D}}$. These rules are mined once offline from the training split and compiled into ASP predicates that can subsequently be reused during inference for an image in $\mathcal{D}$.

\smallskip

\textbf{I}.\hspace{3pt} \textbf{Geometrically-Rooted Spatial Distributions}.\quad For every ordered subject--object pair carrying a predicate $p$, we compute 5 topological relations from the Region Connection Calculus (RCC5) \cite{CohnEtAl1997}, that is, $t\in\{\mathrm{DR,PO,PP,PPI,EQ}\}$ and 9 directions $d\in $ \{ \textit{above}, \textit{below}, \textit{left}, \textit{right}, their intersections, and \textit{same-center}$\}$   \cite{CardinalDirections}. We also compute distinct box bins values for containment, Intersection over Union (IoU), and relative size (6 additional constraints in total).
Conditional empirical distributions $\hat{P}(t\mid p)$, $\hat{P}(d\mid p)$ and their fine-grained triplet variants $\hat{P}(t\mid s,p,o)$, $\hat{P}(d\mid s,p,o)$ are estimated by maximum likelihood. From these, we derive a \emph{forbidden} set of pairs $(p,t)$ with $\hat{P}(t\mid p)\le\tau_{\text{f}}$  used as hard pruning constraints. The same procedure is applied to direction labels. We explain how $\tau_{\text{f}}$ is tuned later.

\smallskip

\textbf{II}.\hspace{3pt}\textbf{Functional Rules.}\quad A predicate $p$ is \emph{functional with degree $N$} if, in any image, at most $N$ distinct subjects relate to a given object through $p$. For each $p$, we sweep the fan-in budget $N\in\{1,2,3,\ldots\}$ and record the empirical violation rate
\begin{equation}\small
  \mathrm{viol}(p,N)\;=\;
  \frac{\#\{(I,o)\,:\,|\{s:(s,p,o)\in I\}|>N\}}
       {\#\{(I,o)\,:\,\exists s:(s,p,o)\in I\}},
\end{equation}

and report the rule confidence $1-\mathrm{viol}(p,N)$. Predicates with $\mathrm{viol}(p,1)=0$ (e.g.\ \emph{wearing}, \emph{riding}) are tagged as \emph{hard}; remaining $(p,N)$ pairs with confidence above $0.95$ are tagged as \emph{soft}. Both are enforced similarly at inference time.

\smallskip

%\textbf{III}.\hspace{3pt}\textbf{Logical Rules.}\quad 

\textbf{III}.\hspace{3pt}\textbf{Relational Propertiess.}\quad We mine three syntactic patterns at the predicate level: (i)~\emph{symmetric}, $p(x,y)\Rightarrow p(y,x)$; (ii)~\emph{inverse}, $p(x,y)\Rightarrow q(y,x)$ for $q\neq p$; and (iii)~\emph{composition}, $p_1(x,y)\wedge p_2(y,z)\Rightarrow p_h(x,z)$. These rules are stored with their support/confidence so that downstream weighted abduction can convert them into log-probability scores.

\smallskip

%\textbf{Output}. 
% \texttt{.lp} file
The above three rule families are emitted as a single ASP program with \emph{spatial}, \emph{functional}, and \emph{logical} sections. Mining is performed once per dataset on the training split only, takes a few minutes on a single CPU, and is decoupled from inference: the rules act purely as data-driven priors that guide constraint-aware filtering.

\smallskip

%\textbf{Rule Application Strategies}. \quad In addition to mining the rules as such, we are also obtaining meta-properties of the rules used for optimisation of the final predicate selection. These properties include the following:

\textbf{Rule Calibration and Deployment.} \quad  In addition to mining commonsense rules from annotated scene graphs, we also estimate a set of rule meta-properties used to control their reliability, granularity, and applicability during inference. These include minimum support thresholds, confidence estimates, predicate- and triplet-level specialization, super-class abstractions, and an empirical verifier stage, as described in the following: 

\smallskip

\textbullet \hspace{4pt}\textbf{Support Threshold}. Depending on the size and diversity of a data source, different minimal support thresholds are applied. The minimal support threshold $\gamma$ represents the minimum number of instances needed to consider as evidence for a rule. We apply a support threshold ($\gamma = 10$) to maintain a reasonable amount of rules generated for big datasets, keeping the solving time within satisfiable bounds.

\smallskip

\textbullet \hspace{4pt}\textbf{Rule Confidence ($\tau$)}
%In addition to a support threshold, we introduce a confidence threshold $\tau$. $\tau$ 
is introduced to fight the annotations noise of SGG datasets 
%. In fact, SGG datasets, especially 
since the ones that are crowd-sourced \cite{krishna2017visual}, can contain wrong/incorrect annotations that will contradict each other. The confidence represent the maximum amount of violation allowed per rule to be validated on the initial data and used for the refinement process. The $\tau$ threshold is tuned during the mining process using a \textit{Verifier} step that test each candidate rule against the predictions obtained by an SGG model on a subset of the data.

\smallskip

\textbullet \hspace{4pt}\textbf{Predicate-Level versus Triplet-Level Rules}. Commonsense rules are mined at two different levels: for a specific predicate (e.g. all relations including \textit{wearing} require $\text{IoU}>0$) or for a specific triplet (e.g. $\langle\text{person},\text{wearing},\text{glasses}\rangle$ requires that the object -- $\textit{glasses}$ -- is smaller than the subject -- $\textit{person}$ -- when the pair is associated with the predicate -- $\textit{wearing}$). Both type of rules are mined similarly, except that $\gamma$ is reduced to $\gamma = 5$ for triplet-level rules.

\smallskip

\textbullet \hspace{4pt}\textbf{Super-Classes}. To address the sparsity of training signal at the fine-grained class level, we introduce a \emph{super-class abstraction} for rule mining.
Each object's category is remapped to a super-class obtained from a pre-computed hierarchical clustering of object classes (at a chosen cut $k$), so that all downstream miners---functional, spatial, and logical rules---operate on pooled super-class triplets (\eg, $\langle$\texttt{super\_car, parked~on, super\_road}$\rangle$) rather than on the hundreds of sparse individual-class triples.
Rules are first mined at the original class resolution; they are then also mined independently on the abstracted data, and the resulting super-class triplet distributions are \emph{expanded} back to every concrete member class and used to fill gaps in the base rule set, with base-level rules always taking precedence.

\smallskip

\textbullet \hspace{4pt}\textbf{Rule Verifier}. Even if each rule is supported by a minimum number of training instances, it can still be the case that some rules are representing noise in the data and will not be beneficial for the refinement process. To address this issue, we introduce a \textit{Verifier} step (see \Cref{fig:approach} Stage 1) during the mining process. The verifier tests each candidate rule against the predictions obtained by an SGG model on a subset of the data. We fire each candidate rule sequentially on the model predictions and evaluate if the rule is switching a True Positive (TP) to a False Positive (FP) or the opposite. Rules that do not demote more TPs than FPs are discarded, while the ones that do are kept for the refinement process. This process is an approximation of the actual impact of each rule on the final performance of our framework, since the ASP-based refinement process is not sequential but rather global and combinatorial. However, it allows us to filter out rules that are not beneficial in a reasonable amount of time.

\subsection{Abductive Scene Graph Refinement}

%We model visual commonsense as a declarative background theory $\Sigma$ expressed in Answer Set Programming (ASP). 

%We model visual commonsense as domain independent structural knowledge denoted by $\Sigma$, consisting of domain-independent structural constraints and inference rules compiled from statistical regularities mined from annotated scene graph datasets. Concretely, these include geometric constraints over spatial configurations, functional constraints over admissible relational cardinalities, and qualitative relational regularities capturing logical dependencies between predicates, implemented as a general set of rules captured in the ASP program $\Pi_{abd}$

We base scene graph refinement on a general abductive reasoning process, where the neural SGG predictions are treated as observations to be explained by an abduced commonsense-consistent scene graph. 
%
%Formally, this is defined as follows: Given the visual commonsense theory $\Sigma$ and the set of neural predictions $\hat{G}$, we abductively infer a scene graph ${G}_{abd}$ such that
%
%\centerline{$\Sigma ~ \wedge ~ {G}_{abd} \models~~ \hat{G} $}
%
%where  ${G}_{abd}$ denotes the set of abductively selected or inferred relations that, together with the commonsense theory $\Sigma$, explain and refine the noisy predictions generated by the neural SGG model.
%
Given the set of neural predictions $\hat{G}$ and the mined commonsense rules, we abductively select or infer a scene graph ${G}_{abd}$ consisting of relations that, together with the commonsense rules, explain and refine the noisy predictions generated by the neural SGG model.
This process is modeled in an abductive reasoning program $\Pi_{\text{abd}}$ consisting of choice rules that generate candidate relations from the relations predicted by the scene graph model, and inference rules and integrity constraints that utilise the mined rules in $\Pi_{\mathcal{D}}$ to restrict these candidate relations.
In particular, these include the following:

\smallskip

\textit{Spatial Constraints} \quad
encode admissible spatial configurations between subject–object pairs. They are represented as constraints over spatial relations derived from bounding boxes (e.g. pertaining to topological relations in RCC5 \cite{CohnEtAl1997}). Forbidden configurations mined from the dataset are compiled into ASP facts such as {\small $\texttt{forbidden\_rcc5\_pred/2}$}, and restrict possible relations based on infeasibility constraints of the following form:
{\small
\begin{align*}
&\texttt{infeasible}(S,P,O) \leftarrow \\
& ~~~~ \texttt{candidate}(S,P,O), \texttt{rcc5\_rel}(S,O,T), \texttt{forbidden\_rcc5\_pred}(P,T).
%& ~~~~ \leftarrow \texttt{candidate}(S,P,O), \texttt{rcc5\_rel}(S,O,T), \texttt{forbidden\_rcc5\_pred}(P,T).
\end{align*}
}

\textit{Functional Constraints} \quad
encode predicate-specific cardinality restrictions across object assignments. Hard-functional predicates mined form the dataset are compiled into predicates {\small $\texttt{functional\_hard\_pred/1}$} and {\small $\texttt{functional\_hard\_trip/3}$}, that are used with ASP integrity constraints, for example to prevent multiple subjects from assigning the same functional relation to a common object, using the following constraint:
{\small
\begin{align*}
&\leftarrow \texttt{functional\_hard\_pred}(P), \texttt{chosen}(S_1,P,O), \texttt{chosen}(S_2,P,O), S_1 \neq S_2.
\end{align*}
}

\textit{Qualitative Relational Regularities} \quad
 encode logical dependencies between predicates and are represented as declarative inference rules. These include symmetric predicates, inverse predicate pairs, relational compositions, incompatibility constraints, and co-occurrence dependencies. For example, relational compositions mined from the dataset are compiled into {\small $\texttt{composition/3}$} facts and instantiated through rules of the form:
{\small
\begin{align*}
&\texttt{candidate}(S,P,O) \leftarrow \\
& ~~~~ \texttt{net\_cand}(S,P_1,M, \_), \texttt{net\_cand}(M,P_2,O, \_), \texttt{composition}(P_1,P_2,P).
\end{align*}
}%
Similarly, incompatibility and symmetry regularities are encoded through ASP rules operating over abductively selected relations.

%This selection process is modeled through ASP choice rules and optimized globally using weighted maximization over candidate scores. 
\textbf{Inference.} \quad 
At inference these ASP specifications are combined with image-specific facts encoded in the ASP program $\Pi_{\mathcal{I}}$, containing the detected objects, their geometric extend, and the predicted relations.
%
%Solving the combined ASP program $\Pi = \Pi_{\mathcal{D}} \cup \Pi_{abd} \cup \Pi_{\mathcal{I}}$ and optimising globally using weighted maximization over candidate scores results in the answer set $\mathbf{SM}$[$\Pi$] consisting of the refined relations. Selected refined relations are inserted as $\hat{\mathcal{T}}_{\text{top-1}}$, and then a new ranking \textit{at the image level} produces the final scene graph.
%
Solving the combined ASP program $\Pi = \Pi_{\mathcal{D}} \cup \Pi_{abd} \cup \Pi_{\mathcal{I}}$ and globally optimising via weighted maximization over the neural confidence scores $f(s,p,o)$ selects the commonsense-consistent answer set $\mathbf{SM}$[$\Pi$] that best preserves the original SGG model confidence. Selected refined relations are inserted as $\hat{\mathcal{T}}_{\text{top-1}}$, and then a new ranking \textit{at the image level} produces the final scene graph.

% \subsubsection{Visual abduction}

% \paragraph{Background knowledge.}
% The dataset-level program $\Pi_\mathcal{D}$ encodes mined constraints as facts of two kinds:
% \[
% \mathtt{forbidden}(p, f). \qquad \mathtt{forbidden\_trip}(c_s, p, c_o, f).
% \]
% where $f$ is any spatial feature value (RCC5 topology, cardinal direction, etc).

% \textcolor{red}{TODO: Add functional, relational rules.}

% \paragraph{Per-image fact base $\Pi_\mathcal{I}$.}
% For each detected object $i$ and candidate pair $(i,j)$:
% \begin{align*}
% &\mathtt{object}(i, c_i).\quad \mathtt{pair}(i,j).\quad \mathtt{feat}(i,j,f_k). \\
% &\mathtt{net\_cand}(i,p,j,v).\quad \mathtt{abd\_cand}(i,p,j,w).
% \end{align*}
% Scores $v,w \in \mathbb{Z}_{>0}$ are integer-scaled network and abduction evidence respectively.

% \paragraph{Complexity.}

\section{Experimental Results}

We evaluate our approach on three standard benchmarks: PSG \cite{yang2022panoptic}, Visual Genome 150 (VG150) \cite{krishna2017visual}, and IndoorVG \cite{neau2023defense}. Visual Genome has become the standard to evaluate SGG approaches. However, it is a noisy dataset with a lot of annotation errors and missing annotations \cite{zellers2018neural,tang2020unbiased}. PSG and IndoorVG are more recent datasets that have been carefully annotated to reduce noise and increase the density of annotations, which makes them more suitable for evaluating the impact of our approach on the consistency of predictions. The PSG dataset is especially interesting because it covers a wide range of contexts and relations, including both spatial and functional ones, which allows us to test the generality of our approach across different types of commonsense constraints.

\subsection{Metrics and Models}

The traditional metrics used in SGG are the Recall@K (R@K) and meanRecall@K (mR@K) \cite{xu2017scene,tang2020unbiased}. 
The Recall@K evaluates the overall performance of a model on the selected dataset whereas meanRecall@K evaluates the performance on the average of all predicate classes, which is more significant for long-tail learning such as in the task of SGG. Both metrics are averaged in the F1@K metric as follows:
$F1@K = \frac{2 \times R@K \times mR@K}{R@K + mR@K}$
which efficiently represents the trade-off for the model performance between head and tail predictions. For both metrics, we evaluate the recall on the top $K$ ($k=[20,50,100]$) relations predicted, ranked by confidence. We also report the Recall@n, where $n$ is the number of ground-truth relations per image.

The Recall@K is a ranking metric which measures only the number of True Positive filtered by a model. However, as previously explained, SGG datasets are extremely sparse and a lot of ground-truth True Positives are not annotated. As a result, our approach cannot be correctly evaluated by only looking at the Recall and meanRecall@K metrics. 

Because our refinement re-assigns predicates to geometrically valid pairs, it can recover \emph{compositionally novel} triplets---subject--predicate--object class combinations absent from the training split---that standard fully-supervised models will struggle to capture. We thus report the zero-shot Recall (zsR@K): R@K computed only over ground-truth triplets whose $(s,p,o)$ class combination never occurs in training.

We additionally introduce the \textbf{Constraint Violation Rate} (CVR): the share of predicted triplets that violate at least one mined constraint. Given a set of spatial constraints $\mathcal{R}$ and predicted triplets $\hat{\mathcal{T}}(I)$ for image $I$,
\begin{equation}\small
  \mathrm{CVR}
    = \frac{\sum_{I} \bigl|\{(s,p,o)\in\hat{\mathcal{T}}(I)
        : \exists\,r\in\mathcal{R},\; r \text{ is violated by } (s,p,o)\}\bigr|}
           {\sum_{I} |\hat{\mathcal{T}}(I)|}.
  \label{eq:cvr}
\end{equation}
%CVR characterises the \emph{raw} inconsistency rate of the network. As the filter explicitly targets these constraints, a reduction is expected by design; we therefore report CVR descriptively---to quantify how inconsistent the baseline is and how much of it the refinement removes---rather than as independent evidence of quality.

%\textbf{Models.} 
We tested our approach with a set of baseline models. We selected models that are representative of the state-of-the-art in SGG, covering a range of architectures and training strategies, to demonstrate the generality of our approach. We used the following models: \numItem{i}~\textbf{Motifs} \cite{zellers2018neural}, a widely used SGG model that incorporates contextual information through a bidirectional LSTM; \numItem{ii}~\textbf{Transformer} \cite{tang2020unbiased}, which leverages self-attention mechanisms to model global context; and \numItem{iii}~\textbf{REACT++} \cite{neau2026react++} which uses cross-attention and prototype learning for real-time scene graph generation. Each of these models was trained on the same datasets using official authors code and evaluated using the same metrics to ensure a fair comparison of the impact of our commonsense refinement approach. For fair comparison we use the same object detection backbone for all models, which is a YOLOv12m \cite{tian2026yolov12} pre-trained on the object detection task of the corresponding dataset.

\subsection{Rules Mining and Commonsense Refinements}
% \subsection{Commonsense Refinement Results} \label{sec:results}

\textbf{Mined Rules.} In \Cref{tab:rule_catalog}, we report the number of rules mined for each family and type, as well as, the number of unique predicate classes covered by these rules. For the PSG dataset, $104,671$ triplet-level, $1,169$ predicate-level and 9 relational rules were mined from the train set. After applying the Verifier step (using the REACT++ model, $\tau=0.09$), $32,665$ rules were retained for the refinement process, which represents 30.9\% of the original rule catalog. 
Symmetry and Inverse are interesting cases, because they are good examples of visual commonsense, however it is very unatural to expect human annotators to annotate both $p(x,y)$ \textit{and} $p(y,x)$ for a symmetric relation, or both $p(x,y)$ \textit{and} $q(y,x)$ for an inverse relation, which makes these rules very difficult to mine from the data. Similarly, compositions are very context-specific and, as a result, they are also hard to mine.

\begin{table}[t]
  \centering
  \caption{\textbf{Rule Catalog}. All mined rules (\emph{base}) vs.\ subset
    selected as beneficial at~$\theta{=}0.15$ (\emph{beneficial}).
    \emph{Constraints}: forbidden (pred./triplet, spatial-relation) entries.
    \emph{Preds}: unique predicate classes covered out of $|V|{=}56$.
  }
  \label{tab:rule_catalog}
  \setlength{\tabcolsep}{5pt}
      \renewcommand{\arraystretch}{0.8}
    \footnotesize
    \sffamily
  \begin{tabular}{@{}llrrrrrr@{}}
  \toprule\rowcolor{headerblue}
    & & \multicolumn{2}{c}{\textbf{Base}} & \multicolumn{2}{c}{\textbf{Beneficial}} & \textbf{Retained} \\
  \cmidrule(lr){3-4}\cmidrule(lr){5-6}
   \cellcolor{headerblue} Group &    \cellcolor{headerblue} Rule type &    \cellcolor{headerblue} Constr. &   \cellcolor{headerblue}  Preds &   \cellcolor{headerblue}  Constr. &    \cellcolor{headerblue} Preds &    \cellcolor{headerblue} (\%) \\
  \midrule
  \cellcolor{headerblue}\multirow{1}{*}{\textbf{Spatial}} & RCC5 (pred.-level) & 161 & 53 & 139 & 53 & 86.3\% \\
  \cellcolor{headerblue}  & RCC5 (triplet-level) & 16,297 & 53 & 10,525 & 53 & 64.6\% \\
  \cmidrule{2-7}
  \cellcolor{headerblue}  & Direction (pred.-level) & 233 & 42 & 214 & 41 & 91.8\% \\
    \cellcolor{headerblue}& Direction (triplet-level) & 6,080 & 34 & 4,065 & 34 & 66.9\% \\
  \cmidrule{2-7}
  \cellcolor{headerblue}  & BBox (pred.-level) & 748 & 53 & 465 & 47 & 62.2\% \\
  \cellcolor{headerblue}  & BBox (triplet-level) & 79,076 & 53 & 14,186 & 53 & 17.9\% \\
  \midrule
  \cellcolor{headerblue}  \multirow{1}{*}{\textbf{Functional}} & Pred.-level & 27 & 27 & 14 & 14 & 51.9\% \\
  \cellcolor{headerblue}  & Triplet-level & 3,218 & 50 & 3,048 & 50 & 94.7\% \\
  \midrule
  \cellcolor{headerblue}  \multirow{1}{*}{\textbf{Relational}} & Composition & 9 & -- & 9 & -- & 100.0\% \\
  \cellcolor{headerblue}  & Symmetry & 0 & -- & 0 & -- & - \\
  \cellcolor{headerblue}  & Inverse  & 0 & -- & 0 & -- & - \\
  \midrule\rowcolor{headerblue}
  \multicolumn{2}{l}{\textbf{Total}} & \textbf{105,829} & & \textbf{32,665} & & \textbf{30.9\%} \\
  \bottomrule
  \end{tabular}
\end{table}

\normalsize

\smallskip

\textbf{Commonsense Refinement.} We present our main results across the three datasets in Table~\ref{tab:main_results}. We compare the baseline model predictions of REACT++ to two variants of our approach: \emph{Base rules} which applies only the rules mined at the original class resolution, and \emph{S-C. \ rules} which additionally applies super-class abstraction. We can observe that our approach consistently improves the F1@K metric across all datasets (from +0.65 to +1.10), with a more significant improvement on the PSG dataset, which is the one with the best annotations quality. The improvement is more modest on the IndoorVG and VG150 datasets, which are noiser, but still significant. We also report the fraction of predicted relations that are reassigned by the filter (Rea.) to show that our approach is able to consistently improving recall by modifying a large portions of predictions (up to 29.6\%). This shows the gap that SGG models still have to fill in terms of performance.

We also observe a consistent improvement with the addition of the super-class abstraction, with relative improvements going from $+0.07$ to $+0.26$ F1@K. This shows that the super-class abstraction is an effective way to address the sparsity of training signal at the fine-grained class level, allowing us to mine more rules and apply them more broadly across the dataset. It is important to notice here that the super-class abstraction can also be applied to \textit{new} object classes that are not present in the training data, as long as they can be mapped to an existing super-class. This can be interesting to extend our approach to Open-Vocabulary or Zero-Shot SGG settings, where the model is expected to generalize to unseen object classes.

\begin{table}[t]
    \centering
    \caption{\textbf{Key Generalization Results Across Datasets}.
    \emph{Base rules} are mined at the original class resolution;
    \emph{S-C.\ rules} additionally apply super-class abstraction ($k{=}12$).
    Each variant uses the threshold $\tau$ maximising F1@K on a held-out sweep ($\tau \in \{0.00, 0.01, \ldots, 0.10\}$).
    Rea.\ = fraction of predicted relations reassigned by the filter.
    Bold denotes the best value per column. Model used: REACT++.}
    \setlength{\tabcolsep}{4pt}
    \renewcommand{\arraystretch}{1.2}
    \footnotesize
    \begin{tabular}{l|l|c|c|c|c|c|c}
        \toprule\rowcolor{headerblue}
        \cellcolor{headerblue}\textbf{Dataset} & \cellcolor{headerblue}\textbf{Rules} & \cellcolor{headerblue}\textbf{Rea.} 
            & \cellcolor{headerblue}\textbf{mR@20/50/100} & \cellcolor{headerblue}\textbf{R@20/50/100}
            & \cellcolor{headerblue}\textbf{R@n} & \cellcolor{headerblue}\textbf{F1@K} & \cellcolor{headerblue}\textbf{$\Delta$F1@K} \\
        \midrule
        \cellcolor{headerblue}\multirow{1}{*}{\textbf{PSG}}
            & None & -- %& 8.39
            & 22.42 / 25.39 / 26.91
            & 31.94 / 37.25 / 40.18
            & 22.41 & 29.59 & --  \\
            \cellcolor{headerblue}& Base \textsubscript{($\theta{=}0.09$)} & 16.5\% % & 1.40
            & 23.34 / 26.34 / 27.58
            & 32.72 / 38.01 / 40.90
            & 22.93 & 30.43 & \textcolor{teal}{+0.84} \\
           \cellcolor{headerblue} & S-C. \textsubscript{($\theta{=}0.10$)} & 18.2\% % & 3.87
            & \textbf{23.66} / \textbf{26.82} / \textbf{27.77}
            & \textbf{33.02} / \textbf{38.27} / \textbf{41.03}
            & \textbf{23.26} & \textbf{30.74} & \textbf{\textcolor{teal}{+1.15}} \\
        \hline
         \cellcolor{headerblue}\multirow{1}{*}{\textbf{IndoorVG}}
            & None & -- % & 19.5
            & 17.54 / 22.52 / 24.89
            & 24.17 / 31.49 / 35.55
            & 13.51 & 25.29 & --  \\
            \cellcolor{headerblue}& Base \textsubscript{($\theta{=}0.09$)} & 28.6\% % & 4.85
            & 17.75 / 23.53 / 25.30
            & 25.75 / 33.12 / 35.98
            & 14.57 & 26.08 & \textcolor{teal}{+0.79} \\
            \cellcolor{headerblue}& S-C. \textsubscript{($\theta{=}0.09$)} & 29.6\% % & 6.87
            & \textbf{17.75} / \textbf{23.55} / \textbf{25.31}
            & \textbf{25.88} / \textbf{33.29} / \textbf{36.23}
            & \textbf{14.59} & \textbf{26.15} & \textbf{\textcolor{teal}{+0.86}} \\
              \hline
         \cellcolor{headerblue}\multirow{1}{*}{\textbf{VG150}}
            & None & -- % & --
            & 10.92 / 14.44 / 16.56
            & 18.96 / 24.54 / 28.11
            & 10.47 & 17.63 & --  \\
            \cellcolor{headerblue}& Base \textsubscript{($\theta{=}0.07$)} & 23.5\% % & --
            & 11.55 / 15.15 / 16.92
            & 20.06 / 25.81 / 28.61
            & 11.08 & 18.34 & \textcolor{teal}{+0.71} \\
            \cellcolor{headerblue}& S-C. \textsubscript{($\theta{=}0.07$)} & 26.7\% % & --
            & \textbf{11.70} / \textbf{15.27} / \textbf{16.92}
            & \textbf{20.38} / \textbf{26.15} / \textbf{28.79}
            & \textbf{11.25} & \textbf{18.49} & \textbf{\textcolor{teal}{+0.87}} \\
        \bottomrule
    \end{tabular}
    \label{tab:main_results}
\end{table}

To demonstrate the generality of our approach across different SGG models, we also applied the same Base rules refinement to the Motifs and Transformer models on the PSG dataset. Results are shown in Table~\ref{tab:models_results}. We can observe that our approach consistently improves the F1@K metric across all models, with relative gains from $+0.25$ to $+0.38$. However, the improvement is more modest compared to the REACT++ model, this can be explained by the lower prediction quality of these models, which leaves less room for improvement.

\smallskip

\textbf{Beyond Standard Recall.} \Cref{tab:zsr} reports two other metrics that provide consistency and quality measure. First, our refinement improves \emph{zero-shot} Recall on all three datasets (e.g.\ zsR@100 $4.31\!\to\!5.21$ on PSG, $+21\%$ relative): by re-assigning predicates to geometrically valid pairs it recovers correct subject--object combinations \emph{never seen in training}---the regime where the base model is weakest. Second, the Constraint Violation Rate of the raw network output is high ($8.5\!-\!12.4\%$) and is sharply reduced by the refinement (below $1\%$ on all three datasets); since the filter targets these constraints this drop is expected, but it quantifies how much latent inconsistency the base predictions contain. In the next section, we detail these results in a series of ablations to better understand the contribution of each type of rules and the interactions between them.

\begin{table}[t]
  \centering
  \caption{\textbf{Additional evaluation} (REACT++): zero-shot Recall@K (zsR@K, $\uparrow$) and Constraint Violation Rate (CVR, $\downarrow$), baseline vs.\ our refinement; all values in \%.}
  \label{tab:zsr}
  \setlength{\tabcolsep}{5pt}\footnotesize\sffamily
      \renewcommand{\arraystretch}{0.7}
  \begin{tabular}{@{}llcccc@{}}
  \toprule\rowcolor{headerblue}
   \cellcolor{headerblue}Dataset & \cellcolor{headerblue} & \textbf{zsR@20} & \textbf{zsR@50} & \textbf{zsR@100} & \textbf{CVR}\,$\downarrow$ \\
  \midrule
  \cellcolor{headerblue}\multirow{1}{*}{VG150} & Baseline & 0.85 & 1.58 & 2.37 & 12.44 \\
   \cellcolor{headerblue}& Ours & \textbf{0.97} & \textbf{1.87} & \textbf{2.74} & \textbf{0.93} \\
  \midrule
  \cellcolor{headerblue}\multirow{1}{*}{IndoorVG} & Baseline & 1.77 & 3.14 & 4.66 & 8.50 \\
   \cellcolor{headerblue}& Ours & \textbf{1.98} & \textbf{3.55} & \textbf{4.79} & \textbf{0.67} \\
  \midrule
  \cellcolor{headerblue}\multirow{1}{*}{PSG} & Baseline & 1.80 & 3.11 & 4.31 & 11.56 \\
   \cellcolor{headerblue}& Ours & \textbf{2.30} & \textbf{3.71} & \textbf{5.21} & \textbf{0.95} \\
  \bottomrule
  \end{tabular}
\end{table}

\begin{table}[t]
    \centering
    \caption{\textbf{Ablation of Models Generalisation on PSG Dataset}.
    Each variant uses the threshold $\theta$ maximising F1@K on a held-out sweep.
    Rea.\ = fraction of predicted relations reassigned by the filter.
    Bold denotes the best value per column.}
    \setlength{\tabcolsep}{1pt}
    \renewcommand{\arraystretch}{1.3}
    \footnotesize
    \sffamily
    \renewcommand{\arraystretch}{0.8}
    \begin{tabular}{>{\columncolor{headerblue}}l|l|c|c|c|c|c|c}
        \toprule\rowcolor{headerblue}
        \textbf{Model} & \textbf{Rules} & \textbf{Rea.}
            & \textbf{mR@20/50/100} & \textbf{R@20/50/100}
            & \textbf{R@n} & \textbf{F1@K} & \textbf{$\Delta$F1@K} \\
        \midrule
        % \multirow{2}{*}{REACT++}
        %     & None (baseline) & --
        %     & 22.42 / 25.39 / 26.91
        %     & 31.94 / 37.25 / 40.18
        %     & 22.41 & 29.59 & -- \\
        %     & Base ($\theta{=}0.06$) & 12.0\%
        %     & \textbf{23.20} / \textbf{26.14} / \textbf{27.45}
        %     & \textbf{32.56} / \textbf{37.84} / \textbf{40.79}
        %     & \textbf{22.75} & \textbf{30.28} & \textbf{\textcolor{teal}{+0.68}} \\
        % \hline
        \multirow{1}{*}{Motifs}
            & None & --
            & 11.60 / 12.92 / 13.56
            & 27.18 / 30.65 / 32.11
            & 19.75 & 17.83 & -- \\

            & Base \textsubscript{($\theta{=}0.10$)} & 13.0\%
            & \textbf{11.76} / \textbf{13.13} / \textbf{13.98}
            & \textbf{27.64} / \textbf{31.16} / \textbf{32.65}
            & \textbf{20.22} & \textbf{18.18} & \textbf{\textcolor{teal}{+0.35}} \\
              \midrule
         \multirow{1}{*}{Transformers}
            & None & --
            & 16.14 / 18.26 / 18.87
            & 28.62 / 32.07 / 33.52
            & 20.76 & 22.68 & -- \\
            & Base \textsubscript{($\theta{=}0.10$)} & 23.5\%
            & \textbf{16.60} / \textbf{18.65} / \textbf{19.35}
            & \textbf{29.11} / \textbf{32.37} / \textbf{33.69}
            & \textbf{21.21} & \textbf{23.13} & \textbf{\textcolor{teal}{+0.45}} \\
        \bottomrule
    \end{tabular}
    \label{tab:models_results}
\end{table}

\subsection{Ablations}

\textbf{Per-rule Attribution.} In this ablation study, we analyze the contribution of each type of rules to the overall improvement of our approach. We focus on the PSG dataset and the REACT++ model, which is the one that shows the most significant improvement. For this model, 685 FP$\to$TP switches are observed on the test set (2,177 images) after applying the commonsense rules refinement. We observe that the majority of these switches are attributed to the spatial rules, either from the rcc5, directionality or box bins categories. Functional and Relational rules are less represented, which can be explained by the fact that they are more context-specific and less frequent in the dataset. Relational rules do not contribute to any switch, in fact Symmetry and Inverse rules are not mined for the PSG dataset, and only 9 Composition rules are mined, which do not fire on any of the test images.

\begin{table}[t]
  \centering
  \caption{\textbf{Per-Rule Type Outcome Analysis}. Each fired rule application is classified by its effect on the GT predicate: FP $\to$ TP (correct promotion),
  TP$\to$FP (erroneous demotion), FP$\to$FP (neutral), TP$\to$TP (preserved).}
  \label{tab:rule_outcomes}
  \setlength{\tabcolsep}{5pt}
        \renewcommand{\arraystretch}{0.8}
  \sffamily\footnotesize
  \begin{tabular}{ll|rrr|rr}
  \toprule\rowcolor{headerblue}
  \textbf{Category} & \textbf{Rule Type} & \textbf{TP $\to$ TP} & \textbf{TP $\to$ FP} & \textbf{FP $\to$ FP} & \textbf{FP $\to$ TP} & \textbf{FP $\to$ TP}, \% \\
  \midrule
  \cellcolor{headerblue} \multirow{1}{*}{Spatial}
  \cellcolor{headerblue}  & RCC5 predicate     &  6 &  85 & 39{,}394 &  93 & 13.6 \\
  \cellcolor{headerblue}  & RCC5 triplet       &  8 &  96 & 37{,}830 &  83 & 12.1 \\
  \cellcolor{headerblue}  & Dir.\ predicate    &  1 &  35 & 26{,}967 &  24 & 3.5 \\
  \cellcolor{headerblue}  & Dir.\ triplet      &  0 &   5 &  8{,}814 &  10 & 1.5 \\
  \cellcolor{headerblue}  & BBox predicate     &  1 &  85 & 43{,}696 &  142 & 20.7 \\
  \cellcolor{headerblue}  & BBox triplet       &  12 & 200 & 60{,}969 & 236 & 34.5 \\
  \midrule
  \cellcolor{headerblue}\multirow{1}{*}{Functional}
  \cellcolor{headerblue} & Predicate          &   0 &   0 &  1{,}284 &  24 & 3.5 \\
  \cellcolor{headerblue}  & Triplet            &   1 &  60 & 29{,}550 &  73 & 10.7 \\
  \midrule
  \cellcolor{headerblue}\multirow{1}{*}{Relational}
  \cellcolor{headerblue}    & Symmetry           &   0 &   0 &        0 &  0 & --- \\
  \cellcolor{headerblue}    & Inverse            &   0 &   0 &        0 &  0 & --- \\
  \cellcolor{headerblue}    & Composition        &   0 &   0 &        0 &  0 & --- \\
  \bottomrule\rowcolor{headerblue}
    & \textbf{Total}     & \textbf{29} & \textbf{566} & \textbf{248{,}504} & \textbf{685} & \textbf{100.0} \\
  \bottomrule
  \end{tabular}
\end{table}

\smallskip

\begin{figure}[t]
  \centering
  \begin{minipage}[t]{0.47\linewidth}
    \centering
    \includegraphics[width=\linewidth]{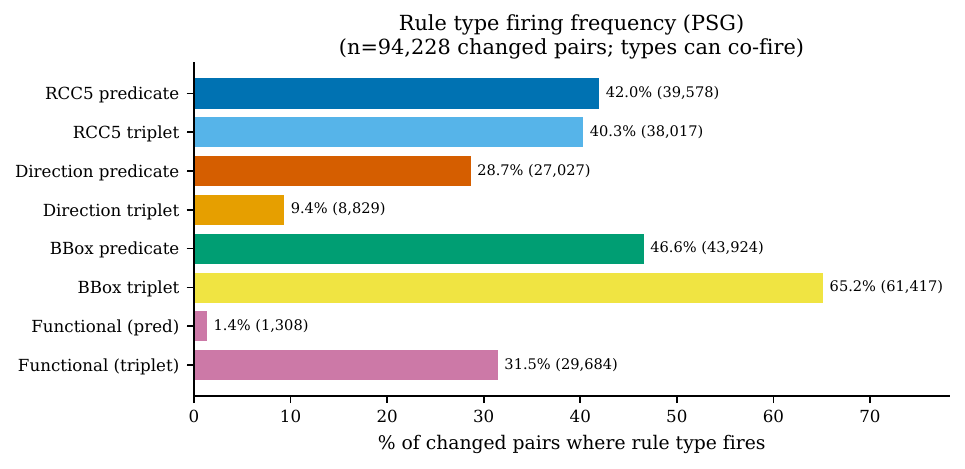}
  \end{minipage}\hfill
  \begin{minipage}[t]{0.47\linewidth}
    \centering
    \includegraphics[width=\linewidth]{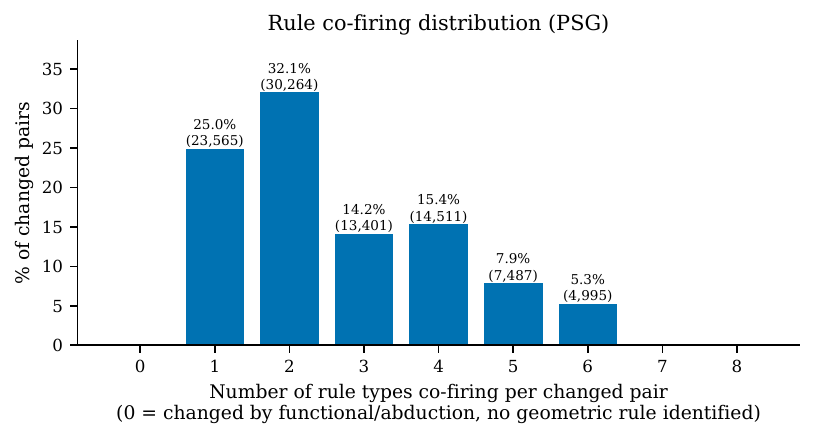}
  \end{minipage}
  \caption{\textbf{Rule Firing on PSG}. \emph{Left:} firing frequency of each rule
  type, expressed as the percentage of changed pairs in which a rule of that
  type fires (types can co-fire, so percentages sum to more than $100$).
  \emph{Right:} distribution of the number of rule types co-firing per
  changed pair; roughly half of all changed pairs trigger three or more rule
  types simultaneously.}
  \label{fig:firing}
\end{figure}

\textbf{Co-firing Rule Combinations.} \Cref{fig:firing}~(left) summarises rule-type firing frequency (PSG dataset, REACT++ model). Across the 94,228 pairs whose predicate changes under the filter, bounding-box features (predicate and triplet level combined) fire on roughly two thirds of changed pairs, RCC5 on about $40\%$, direction on about $30\%$, and functional on a third. Pred-level and triplet-level rules can co-fire on the same pair, so the percentages sum to more than $100$. The co-firing distribution (\Cref{fig:firing}, right) confirms that this overlap is the rule rather than the exception: $32.1\%$ of changed pairs trigger exactly two rule types and $42.8\%$ trigger three or more, with only a quarter of changes driven by a single rule type.
 
% \begin{table}[h]
%   \centering
%   \caption{\textbf{Co-Firing Rule-Types}. Most frequently co-firing combinations among the 171 FP$\to$TP switches.  Each row is one unique attribution set.}
%   \label{tab:rule_cofiring}
%   \sffamily
%   \footnotesize
%   \begin{tabular}{l|r|r}
%     \toprule\rowcolor{headerblue}
%     \textbf{Rule-Type Combination} & \textbf{Count} & \textbf{\%} \\
%     \midrule
%     \textsc{bbox} trip only                                & 81 & 47.4 \\
%     \textsc{rcc5} pred $+$ \textsc{bbox} trip              & 51 & 29.8 \\
%     \textsc{rcc5} pred $+$ \textsc{rcc5} trip $+$ \textsc{bbox} trip & 21 & 12.3 \\
%     \textsc{rcc5} pred $+$ \textsc{rcc5} trip $+$ \textsc{bbox} trip $+$ func.\ hard $+$ func.\ soft & 9 & 5.3 \\
%     \textsc{rcc5} pred only                                &  6 &  3.5 \\
%     Direction pred $+$ direction trip $+$ \textsc{bbox} trip &  3 &  1.8 \\
%     \bottomrule
%   \end{tabular}
% \end{table}

\begin{figure}
    \centering
    \includegraphics[width=0.47\linewidth]{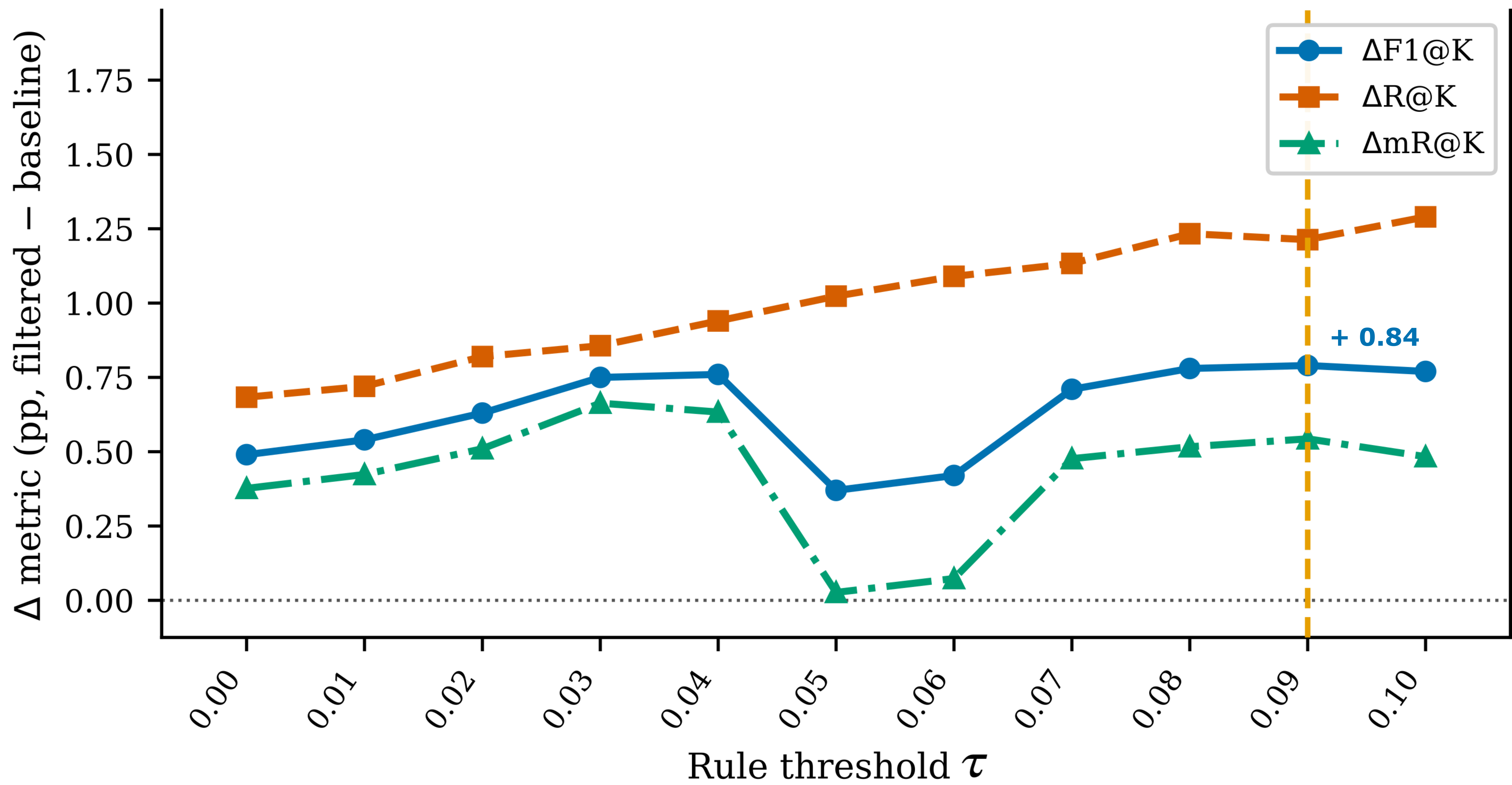}
    \caption{\textbf{{\large$\tau$} Tuning}. Effect of varying the confidence threshold $\tau$ on PSG performance metrics (REACT++ model, PSG dataset).}
    \label{fig:thresholds_react_psg}
\end{figure}

\textbf{{\large$\tau$} Tuning.} \quad In \Cref{fig:thresholds_react_psg}, we show the effect of varying the threshold $\tau$ on the PSG performance metrics. With $\tau=0.0$, we do not authorize any violation of the mined rules, which is a strict setting that assume that all annotations in the training data are correct and that all mined rules are perfectly valid. As $\tau$ increases, we allow more violations of the mined rules, which can be beneficial to account for noise in the data and to avoid overfitting to the training set. We can observe that there is a sweet spot around $\tau=0.09$ where the F1@K metric is maximized, which indicates that allowing a small amount of violation can lead to better generalization on the test set.

\begin{wraptable}{r}{0.5\textwidth}   % {placement}{block width}
  \centering
\centering
\setlength{\tabcolsep}{2pt}
      \renewcommand{\arraystretch}{0.6}
\sffamily
\footnotesize
\vspace{-10pt}
\begin{tabular}{lrrrrr}
\toprule\rowcolor{headerblue}
\textbf{ASP / Clingo {\normalfont\citep{Gebser2016-clingo5}} Phase} &  \textbf{Mean} & \textbf{Median} & \textbf{p95}  \\
\midrule
% Python: build LP / facts        &   2.1 &   1.3 &   4.4  \\
% Clingo: load background LP      & 113.4 & 113.2 & 115.4  \\
%add facts               &   6.6 &   4.4 &  14.5  \\
\cellcolor{headerblue}Grounding                  &  12.6 &   9.3 &  25.0  \\
\cellcolor{headerblue}Solving (Computing Models)                   &  84.2 &   0.5 &   2.0  \\
\midrule\rowcolor{headerblue}
\textbf{Total per image}        & 96.8 & 9.8 & 27.0  \\
\bottomrule
\end{tabular}
\normalfont\footnotesize
\caption{\textbf{Per-Image ASP-Based Reasoning}. Runtime Performance for PSG in Milliseconds.  Benchmarked using an Intel Core Ultra 5 245 x 14 CPU, 32GB RAM.}
\label{tab:asp-timing}
\end{wraptable}
\textbf{Compute Time.} \Cref{tab:asp-timing} presents runtime performance of the ASP refinement process on the PSG dataset, with the same rule catalog as in \Cref{tab:rule_catalog} (32,665 rules). The total time per image is 96.8 ms on average, with the majority of the time spent in the Clingo \citep{Gebser2016-clingo5} solving phase (84.2 ms). This time is inflated a lot by the functional rules, which can take up to a second to solve on complex images. However, since only a small fraction of images trigger functional rules, the median time for solving is much lower (0.5 ms). The grounding phase takes 12.6 ms on average. Overall, the ASP refinement process is efficient enough to be applied in real-time settings, especially if we consider that it can be parallelized across images under different threads.

\subsection{Qualitative Analyses of Results}

%\begin{figure}
%    \centering
%    \includegraphics[width=0.85\linewidth]{figures/qualitative_ex2.png}
%    \caption{\textbf{Qualitative Example for a Functional Rule} - PSG dataset, REACT++ model.}
%    \label{fig:qualitative_ex1}
%\end{figure}
\begin{figure}
    \centering
    \includegraphics[width=0.96\linewidth]{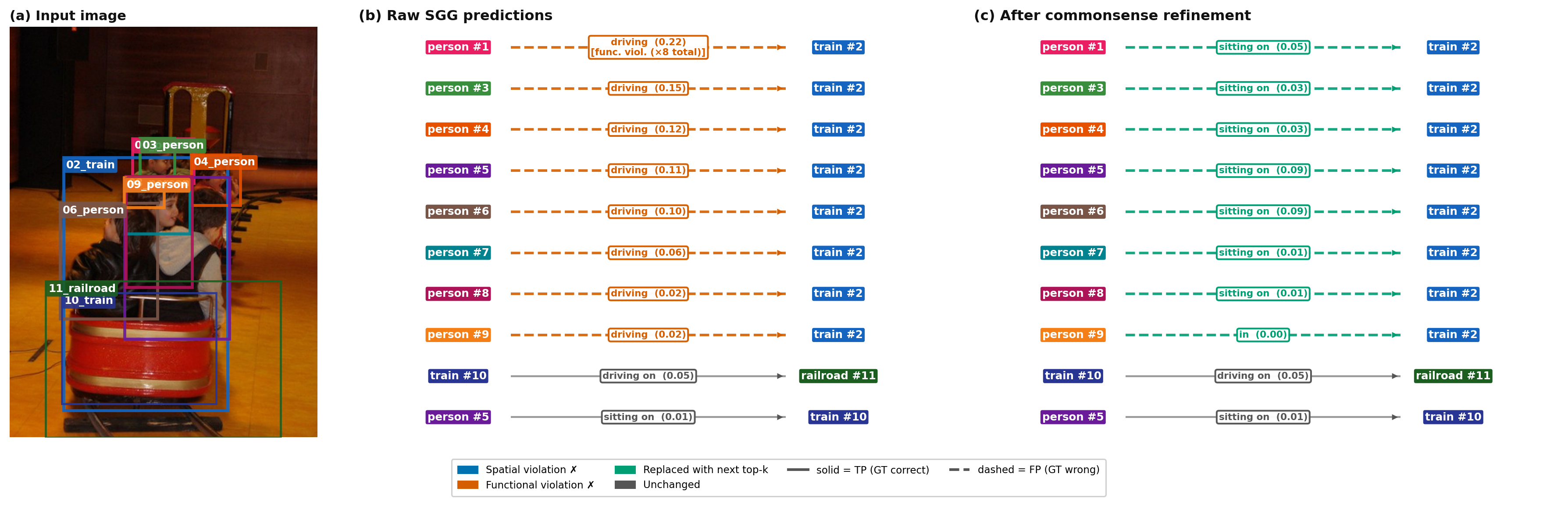}\\[4pt]
    \includegraphics[width=0.96\linewidth]{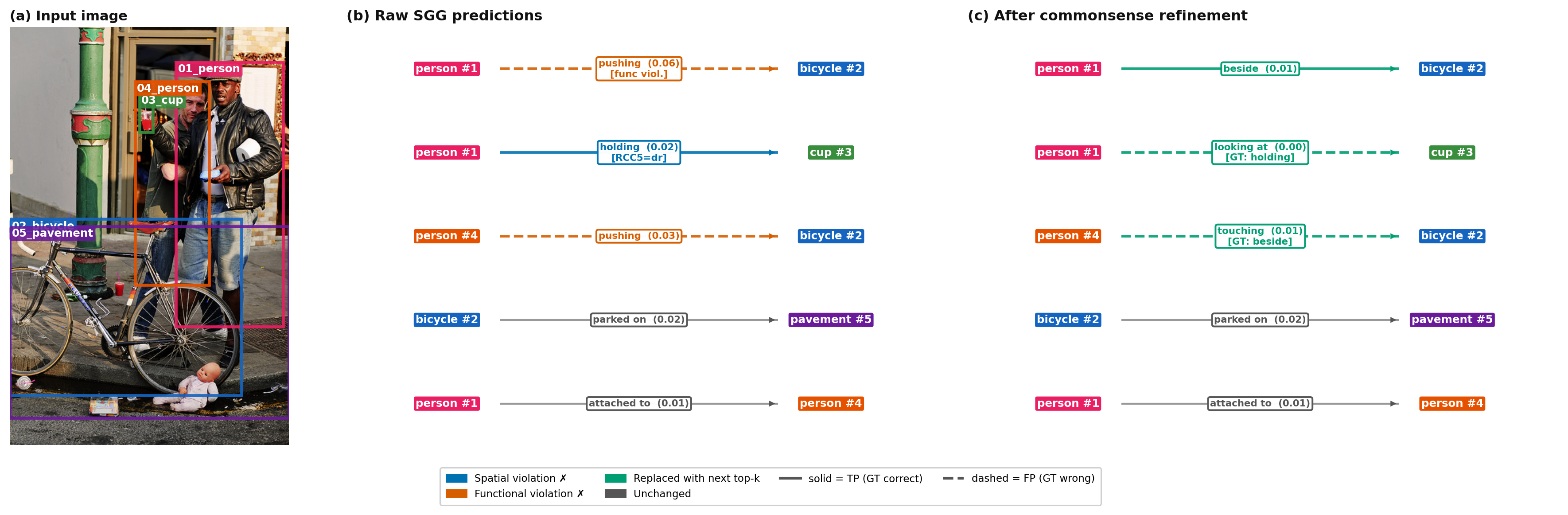}
    \caption{\textbf{Qualitative Functional and Spatial Rules} - PSG dataset, REACT++ model.}
    \label{fig:qualitative_ex1}
\end{figure}
In this section, we present empirical evidence of our approach, as well as illustration of limitations. 
In \Cref{fig:qualitative_ex1} (top) we consider the case of a functional violation with the \textit{driving} predicate (functional - predicate level). The rule states that a given object can be related to at most one subject through the predicate \textit{driving}. In this example, the initial predictions of the SGG model (left) contain 8 \textit{driving} relations for the same object (\texttt{train}), which violates the functional constraint. The ASP-based refinement process identifies this violation and reassigns all invalid relations to more plausible predicates, such as \textit{sitting on} (right). This re-assignment is not increasing the recall on this image because the GT annotations do not contain the \textit{sitting on} relation, but it is a correct decision given the visual evidence. This example shows the limitations of recall-based metrics.
In \Cref{fig:qualitative_ex1} (bottom) we consider the case of the relation \textit{holding} a \textit{cup}. The spatial rule states that a \textit{holding} predicate should respect an intersection between subject and object (IoU > 0.0). In this example, we can see that the \textit{person 1} bounding box is not intersecting with the \textit{cup} bounding box, which violates the spatial constraint. The confusion stems from the fact that the person bounding box is not well annotated and does not contain the hand of the person, which is the actual subject of the \textit{holding} relation. The ASP-based refinement is then demoting the True Positive to a False Positive, which is a correct decision given the annotation, but gives a negative reward in the recall-based metrics. This example highlights the fact that the annotations are not perfect and that our method can also have a negative impact when data quality is low.

% \begin{figure}
%     \centering
%     \includegraphics[width=0.5\linewidth]{figures/cardinality_cmp_2352307.png}
%     \caption{\textbf{Qualitative Example for a Functional Rule}. Consider the case of \textit{driving}. The rule states that a given object can be related to at most one subject through the predicate \textit{driving}. In this example, the initial predictions of the SGG model (left) contain 15 \textit{driving} relations for the same object (\texttt{train}), which violates the functional constraint. The ASP-based refinement process identifies this violation and reassigns all invalid relations to more plausible predicates, such as \textit{sitting on} (right).}
%     \label{fig:qualitative}
% \end{figure}

\section{Conclusion and Outlook}

\emph{We propose} a systematic method wherein visual commonsense, encompassing spatial feasibility, role-cardinality, and logical entailment regularities governing visual relations, are enforced by an explicit (declarative) reasoning stage rather being recovered through more capacity or data.  The (neurosymbolic) synergy we implement is a principled and specialised integrated of diverse capacities: neural visual computing methods are neither designed nor trained to discover commonsense rules. Driven by this, our method mines spatial, functional, and relational constraints directly from training annotations and utilised them in a declarative commonsense reasoning stage (in answer set programming) that prunes infeasible predictions and abduces missing scene graph knowledge, with a verifier retaining only the empirically beneficial subset. Unlike knowledge-infusion or LLM-oracle refinement methods, the constraints are dataset-aligned, transparently grounded in detected geometry, and formally verifiable and human explainable for every correction, which matters because consistency and robustness, not raw ranking, are the principal concerns within application domains. 

\emph{As outlook}, we aim to further improve empirical outcomes, particularly concerning recall-based metrics. This is motivated by the fact that these benchmarks do not account for the fact that many fitting annotations remain unaccounted; for instance, symmetry and inverse patterns are almost impossible to mine, since annotators typically do not label both directions of such relations. This is less a weakness of our method than evidence that recall mismeasures consistency-oriented refinement, motivating evaluation metrics --beyond our Constraint Violation Rate-- that credit logically entailed and feasibility-corrected edges directly. Additionally, we also intend to extend the mined commonsense from spatial feasibility toward functional affordances and contextual regularities through the incorporation of diverse forms of formalised and generative domain-dependent and independent commonsense knowledge. 
%Finally, we aim to extend the declarative reasoning to include weighted soft constraints, which can be used to further improve recall by allowing the recovery of more uncertain but still plausible relations.

\newpage

\bibliography{bibs/sgg_commonsense,bibs/ASP-VLM}

\clearpage
\appendix

\section*{Contents}

This supplement provides additional material supporting including the declarative encoding used at inference time, a complete
catalogue of the commonsense rule families mined from the training data, an
empirical analysis of how the filter behaves on PSG, and extended figures. The following are included herein:

\begin{enumerate}
  \item \textbf{Declarative Refinement Encoding} 
  
  The two-stage Answer Set
        Program (ASP) encoding: dataset-level background facts, per-image
        facts, the stage-1 spatial filter, and the stage-2 functional
        constraint program.
  
  \item \textbf{Mined Rule Catalogue.} 
  
  The complete set of mined rules: topology, cardinal direction, bounding-box features, functional
        fan-in, and relational properties.
        
  \item \textbf{Filter Behaviour on PSG.} 
  
  Empirical analysis of which rules
        fire on the test split, how they affect ground-truth predictions, and
        their aggregate F1 contribution.
        
  \item \textbf{Extended Figures.} 
  
  Clustering dendrograms, geometric
        statistics, per-relation RCC5 analyses, and additional per-predicate
        plots.
\end{enumerate}

\newpage

\section{Declarative Refinement Encoding}
\label{app:asp}

The commonsense reasoning stage is realised as a declarative program in Answer
Set Programming (ASP) and solved with Clingo, one program per image. The
pipeline runs in two stages. \emph{Stage~1} (Section~\ref{sec:stage1}) takes
the network's ranked per-pair predictions together with the mined geometric
constraints and selects, for each pair independently, the highest-scoring
predicate that is consistent with the mined catalogue. \emph{Stage~2}
(Section~\ref{sec:functional}) enforces cross-pair functional constraints
(fan-in equal to one) by re-selecting the predicate assignment over the
conflict clusters that remain after stage~1.

\subsection{Background Facts}

The mined rule catalogue (Section~\ref{app:rules}) is compiled once per
dataset into a set of background facts. A \texttt{forbidden\_*} fact states
that a predicate must not hold when the subject--object pair exhibits a given
geometric relation. Each constraint is available at predicate level and at the
finer-grained $\langle$subject class, predicate, object class$\rangle$ triplet
level. Relational-property composition is compiled into
\texttt{horn\_composition/3} facts that bridge two network candidates into a
third (Section~\ref{sec:relprop}). For each image, the detector outputs and the network's predicate distribution
are emitted as facts. The measured geometric relation
of every candidate pair is computed from the two bounding boxes and emitted as
\texttt{rcc5\_rel/3}, \texttt{dir\_rel/3}, and one \texttt{bbox\_feat/3} fact
per discretised feature.

\begin{lstlisting}[style=prolog]
%% RCC5 topology
forbidden_rcc5_pred(P, T).            % predicate-level
forbidden_rcc5_trip(SC, P, OC, T).    % triplet-level

%% Cardinal direction
forbidden_dir_pred(P, D).
forbidden_dir_trip(SC, P, OC, D).

%% Discretised bounding-box feature bin
forbidden_bbox_pred(P, F).
forbidden_bbox_trip(SC, P, OC, F).

%% Relational-property composition
horn_composition(P1, P2, Ph).
\end{lstlisting}

\begin{lstlisting}[style=prolog]
object(O, C).            % object O has class C
pair(S, O).              % ordered candidate pair: subject S, object O
rcc5_rel(S, O, T).       % measured RCC5 topology between S and O
dir_rel(S, O, D).        % measured cardinal direction
bbox_feat(S, O, F).      % one fact per discretised bbox-feature bin
net_cand(S, P, O, V).    % network candidate: predicate P, integer score V
abd_cand(S, P, O, W).    % spatial-support candidate: predicate P, score W
\end{lstlisting}

\subsection{Stage 1: Spatial Filtering and per-pair selection}
\label{sec:stage1}
A candidate is ruled out when its measured spatial relation is forbidden for 
its predicate, either at predicate level or at triplet level. Relational-property composition adds a
candidate whenever two network candidates form the body of a mined
composition rule. For each pair, the highest-scoring feasible network
candidate is selected, with the predicate atom order breaking ties; a
spatial-support fallback selects a candidate only for pairs where every
network candidate has been ruled out:

\newpage

\begin{lstlisting}[style=prolog]
%% 1. Candidates
candidate(S,P,O) :- net_cand(S,P,O,_).
candidate(S,P,O) :- abd_cand(S,P,O,_).

%% Relational-property composition
candidate(S,P,O) :- net_cand(S,P1,M,_), net_cand(M,P2,O,_),
                    horn_composition(P1,P2,P).

%% 2. Infeasibility
infeasible(S,P,O) :- candidate(S,P,O), rcc5_rel(S,O,T),
                     forbidden_rcc5_pred(P,T).

infeasible(S,P,O) :- candidate(S,P,O), rcc5_rel(S,O,T),
                     object(S,SC), object(O,OC),
                     forbidden_rcc5_trip(SC,P,OC,T).

infeasible(S,P,O) :- candidate(S,P,O), dir_rel(S,O,D),
                     forbidden_dir_pred(P,D).

infeasible(S,P,O) :- candidate(S,P,O), dir_rel(S,O,D),
                     object(S,SC), object(O,OC),
                     forbidden_dir_trip(SC,P,OC,D).

infeasible(S,P,O) :- candidate(S,P,O), bbox_feat(S,O,F),
                     forbidden_bbox_pred(P,F).

infeasible(S,P,O) :- candidate(S,P,O), bbox_feat(S,O,F),
                     object(S,SC), object(O,OC),
                     forbidden_bbox_trip(SC,P,OC,F).

%% 3. Feasible candidates
feasible(S,P,O) :- candidate(S,P,O), not infeasible(S,P,O).

%% 4. Network-first selection
has_net_feasible(S,O) :- feasible(S,P,O), net_cand(S,P,O,_).

best_net_score(S,O,W) :- has_net_feasible(S,O),
                         W = #max { V : net_cand(S,P,O,V), feasible(S,P,O) }.

chosen(S,P,O) :- best_net_score(S,O,W),
                 P = #min { P2 : net_cand(S,P2,O,W), feasible(S,P2,O) }.

%% 5. Spatial-support fallback: only for pairs with no feasible net candidate
has_abd_feasible(S,O) :- feasible(S,P,O), abd_cand(S,P,O,_).

best_abd_score(S,O,W) :- has_abd_feasible(S,O), not has_net_feasible(S,O),
                         W = #max { V : abd_cand(S,P,O,V), feasible(S,P,O) }.

chosen(S,P,O) :- best_abd_score(S,O,W), not has_net_feasible(S,O),
                 P = #min { P2 : abd_cand(S,P2,O,W), feasible(S,P2,O) }.

#show chosen/3.
\end{lstlisting}

\subsection{Stage 2: Functional constraints}
\label{sec:functional}

Stage~1 selects a predicate for each pair independently. Cross-pair coupling
remains: an object should not receive the same \emph{hard-functional}
predicate (\eg{} \emph{feeding}, \emph{guiding}, \emph{throwing}) from more
than one subject in the same image. Stage~2 runs only when stage~1
produces such a conflict; the stage-1 output and each pair's top-$K$
feasible alternates are passed to a small per-image program whose choice
rule selects at most one predicate per pair, integrity constraints enforce
hard functional fan-in, and \texttt{\#maximize} optimises the total stage-1
score.

\begin{lstlisting}[style=prolog]
#defined cand/4.
#defined object/2.
#defined functional_hard_pred/1.
#defined functional_hard_trip/3.

pair(S,O) :- cand(S,_,O,_).

{ chosen(S,P,O) : cand(S,P,O,_) } <= 1 :- pair(S,O).

:- functional_hard_pred(P), chosen(S1,P,O), chosen(S2,P,O), S1 != S2.

:- functional_hard_trip(SC,P,OC),
   chosen(S1,P,O), chosen(S2,P,O), S1 != S2,
   object(S1,SC), object(S2,SC), object(O,OC).

#maximize { W,S,P,O : chosen(S,P,O), cand(S,P,O,W) }.

#show chosen/3.
\end{lstlisting}

\subsection{Minimal (Worked) Example}
\label{sec:worked}

\begin{figure}[h]
  \centering
  \includegraphics[width=0.5\linewidth]{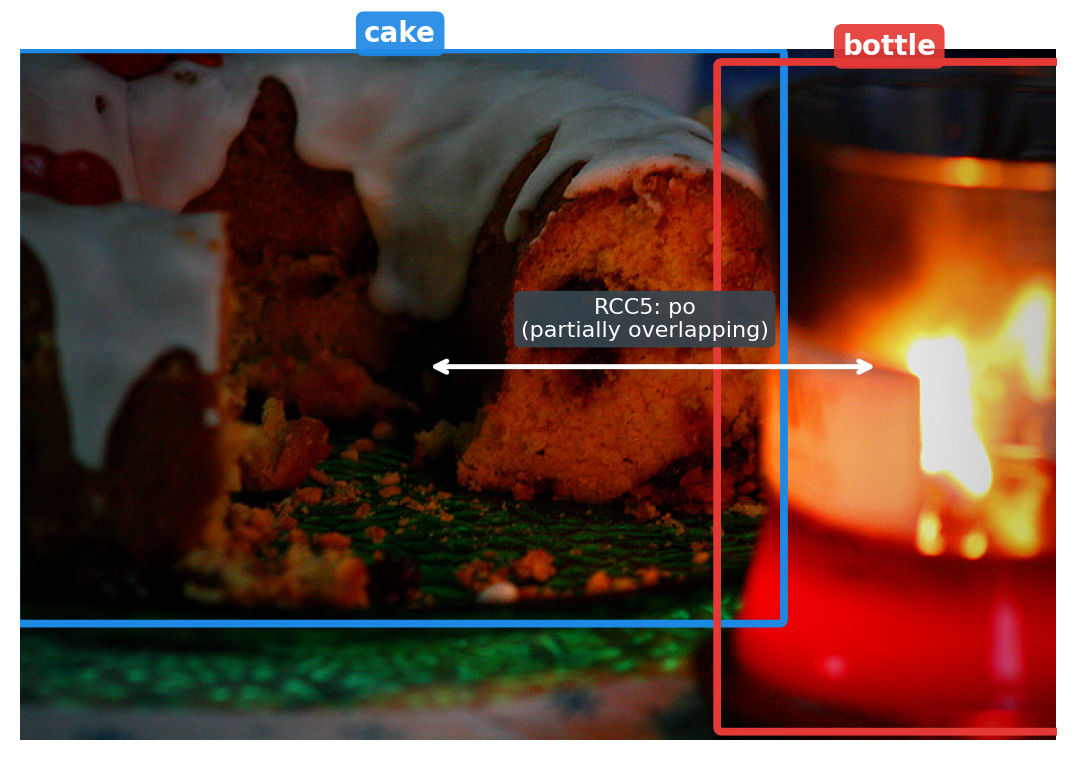}
  \caption{PSG test image \texttt{2316458}: a bottle and a cake in partial
  overlap. The two bounding boxes share interior pixels but neither
  contains the other, yielding RCC5 topology \texttt{po}.}
  \label{fig:worked-example}
\end{figure}

To illustrate the encoding on a concrete instance, we show the stage-1
solve for PSG test image \texttt{2316458} (Figure~\ref{fig:worked-example}),
in which two objects are detected: a bottle and a cake placed side by side.
The two pairs are in partial overlap (\texttt{po}), so
no \texttt{dir\_rel} fact is emitted (cf.\ Section~\ref{sec:relprop}). 
The network's top-1 prediction for both directions is already
\emph{beside}, and the filter confirms it. The value added is twofold:
first, the 20 \texttt{infeasible/3} atoms are predicates that the mined
geometric calculi rule out a priori for this bounding-box configuration;
had the network ranked any of these above \emph{beside}, the filter would
have demoted them without supervision. Second, the program leaves an
auditable trail of exactly which constraints fired -- a property no
opaque post-hoc re-ranker provides. The network's top-1 prediction for both directions is already
\emph{beside}, and the filter confirms it. The value added is twofold:
first, the 20 \texttt{infeasible/3} atoms are predicates that the mined
geometric calculi rule out a priori for this bounding-box configuration;
had the network ranked any of these above \emph{beside}, the filter would
have demoted them without supervision.
\begin{lstlisting}[style=prolog]
% Objects
object(bottle_1360, bottle).
object(cake_1359,   cake).

% Measured RCC5 topology
rcc5_rel(bottle_1360, cake_1359,   po).
rcc5_rel(cake_1359,   bottle_1360, po).

% Top-5 network candidates per direction
net_cand(bottle_1360, beside,       cake_1359,   74).
net_cand(bottle_1360, attached_to,  cake_1359,   15).
net_cand(bottle_1360, in_front_of,  cake_1359,    6).
net_cand(bottle_1360, on,           cake_1359,    5).
net_cand(bottle_1360, in,           cake_1359,    3).
net_cand(cake_1359,   beside,       bottle_1360, 56).
net_cand(cake_1359,   attached_to,  bottle_1360, 20).
net_cand(cake_1359,   in_front_of,  bottle_1360,  7).
% ...
\end{lstlisting}

\begin{lstlisting}[style=prolog]
% Predicates ruled out by mined geometric constraints (20 atoms)
infeasible(bottle_1360, about_to_hit,  cake_1359).
infeasible(bottle_1360, catching,      cake_1359).
infeasible(bottle_1360, climbing,      cake_1359).
infeasible(bottle_1360, cooking,       cake_1359).
infeasible(bottle_1360, entering,      cake_1359).
infeasible(bottle_1360, jumping_from,  cake_1359).
infeasible(bottle_1360, jumping_over,  cake_1359).
infeasible(bottle_1360, kicking,       cake_1359).
infeasible(bottle_1360, painted_on,    cake_1359).
infeasible(bottle_1360, riding,        cake_1359).
infeasible(bottle_1360, swinging,      cake_1359).
infeasible(bottle_1360, throwing,      cake_1359).
infeasible(cake_1359,   about_to_hit,  bottle_1360).
infeasible(cake_1359,   catching,      bottle_1360).
infeasible(cake_1359,   climbing,      bottle_1360).
infeasible(cake_1359,   entering,      bottle_1360).
infeasible(cake_1359,   kicking,       bottle_1360).
infeasible(cake_1359,   painted_on,    bottle_1360).
infeasible(cake_1359,   swinging,      bottle_1360).
infeasible(cake_1359,   throwing,      bottle_1360).

% Final per-pair selection
chosen(bottle_1360, beside, cake_1359).
chosen(cake_1359,   beside, bottle_1360).
\end{lstlisting}

\section{Mined Rule Catalogue}
\label{app:rules}

From the training annotations we mine three families of commonsense rules:
\emph{spatial} regularities (RCC5 topology, cardinal direction, and
bounding-box features), \emph{functional} constraints, and \emph{relational
properties}. Mining is performed once per dataset on the training split, using
a minimum support of $\gamma=10$ instances at the predicate level and
$\gamma=5$ at the triplet level. This section documents each family in turn.

\subsection{RCC5 topology}

\begin{figure}[t]
  \centering
  \includegraphics[width=0.75\linewidth]{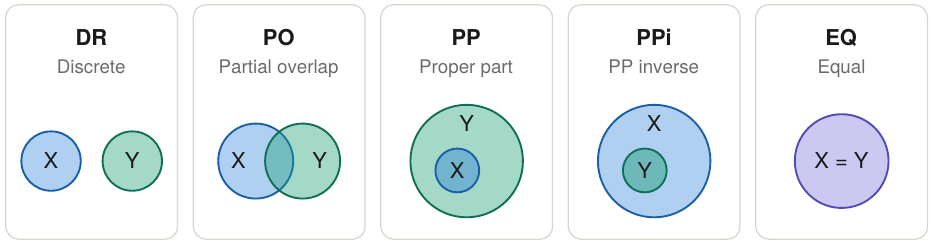}
  \caption{2D Interpretations of the RCC5 Topological Relations.}
  \label{fig:rcc5}
\end{figure}

\begin{figure}[htbp]
  \centering
  \includegraphics[width=0.95\linewidth]{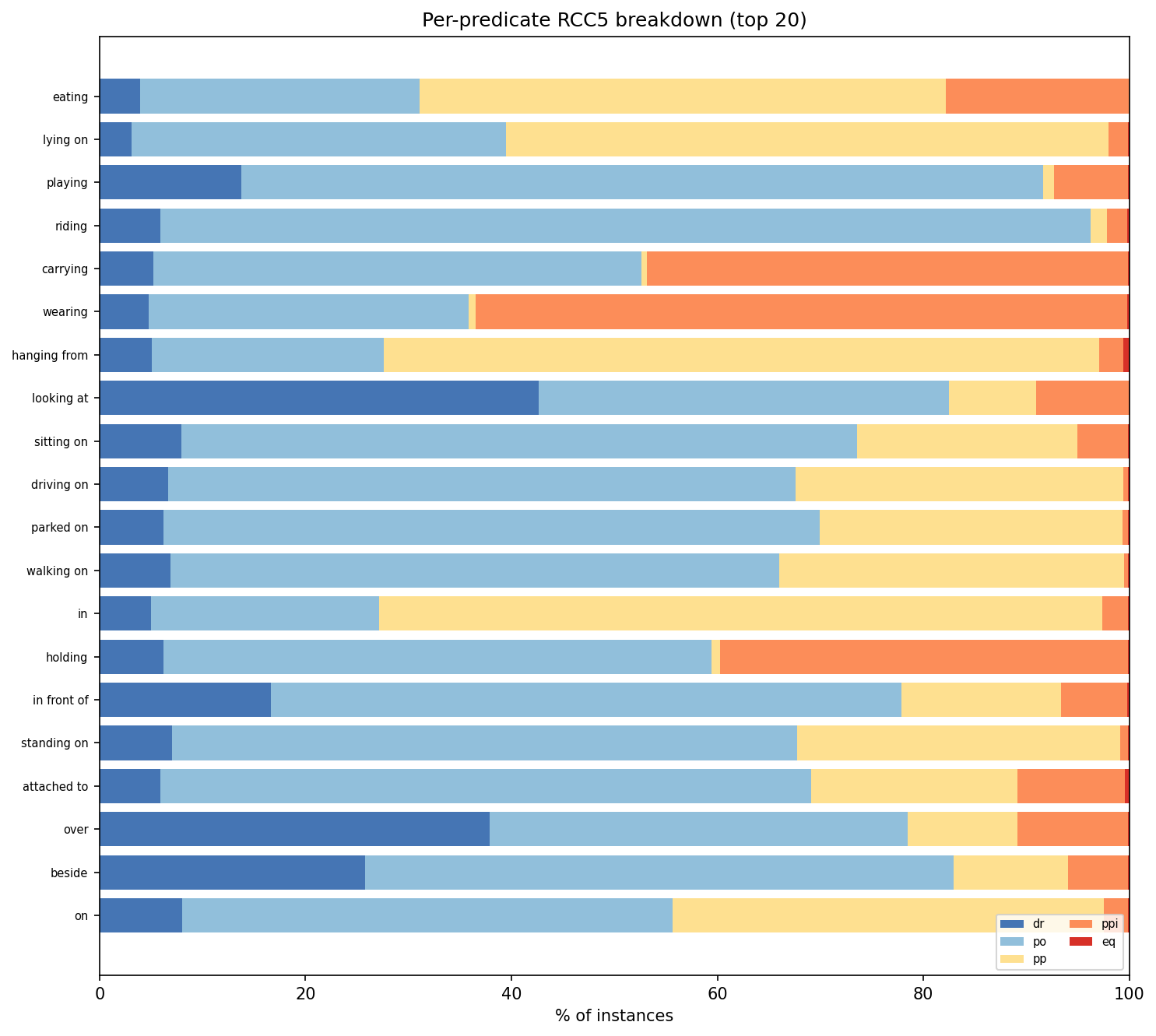}
  \caption{Per-predicate distribution over the five RCC5 topologies for PSG.}
  \label{fig:rcc5-stacked}
\end{figure}

\begin{figure}[htbp]
  \centering
  \includegraphics[width=0.95\linewidth]{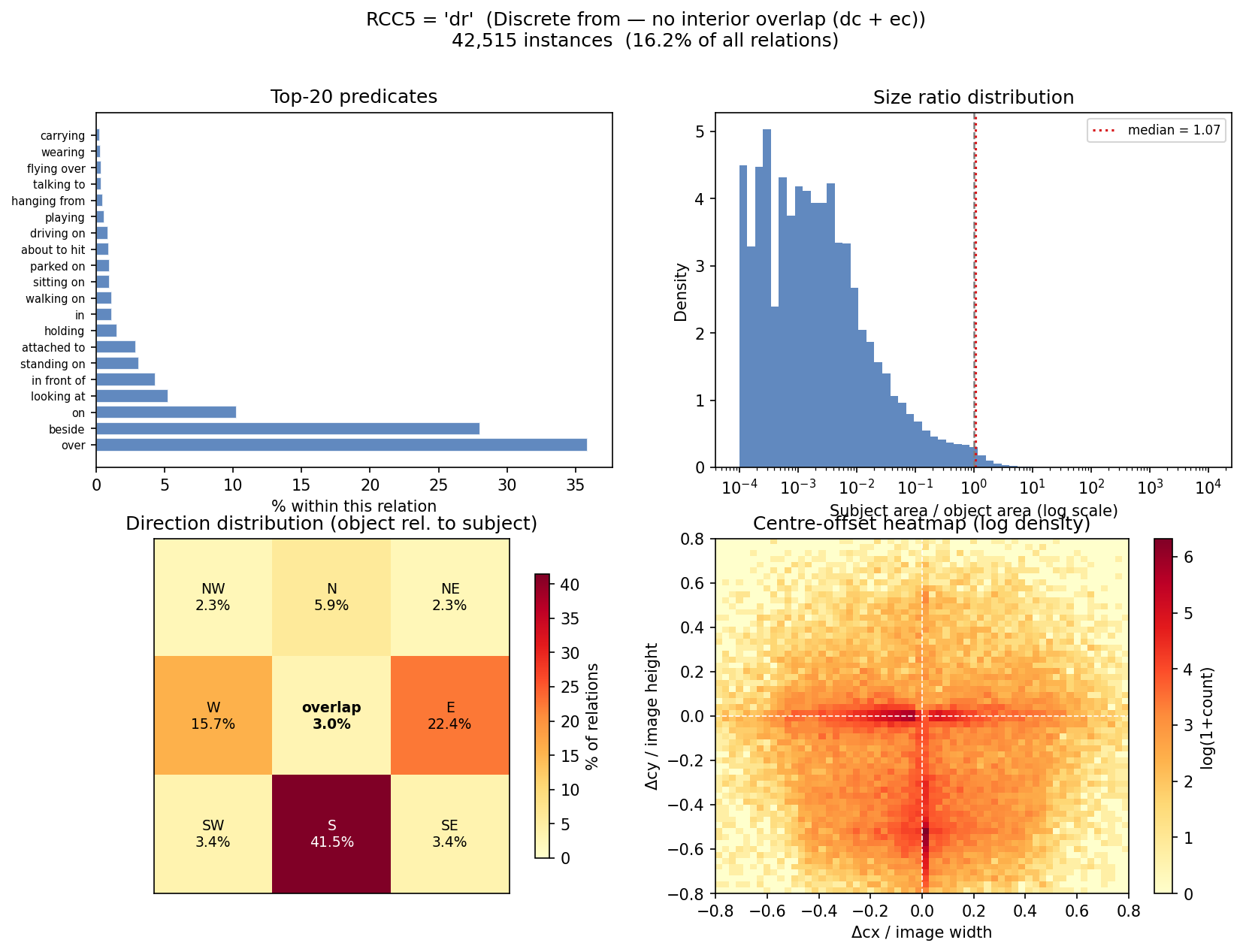}
  \caption{PSG predicates associated with the RCC5 \emph{discrete}
  (\texttt{dr}) topology -- the principal no-overlap case.}
  \label{fig:rcc5-dr}
\end{figure}

\begin{figure}[htbp]
  \centering
  \includegraphics[width=0.95\linewidth]{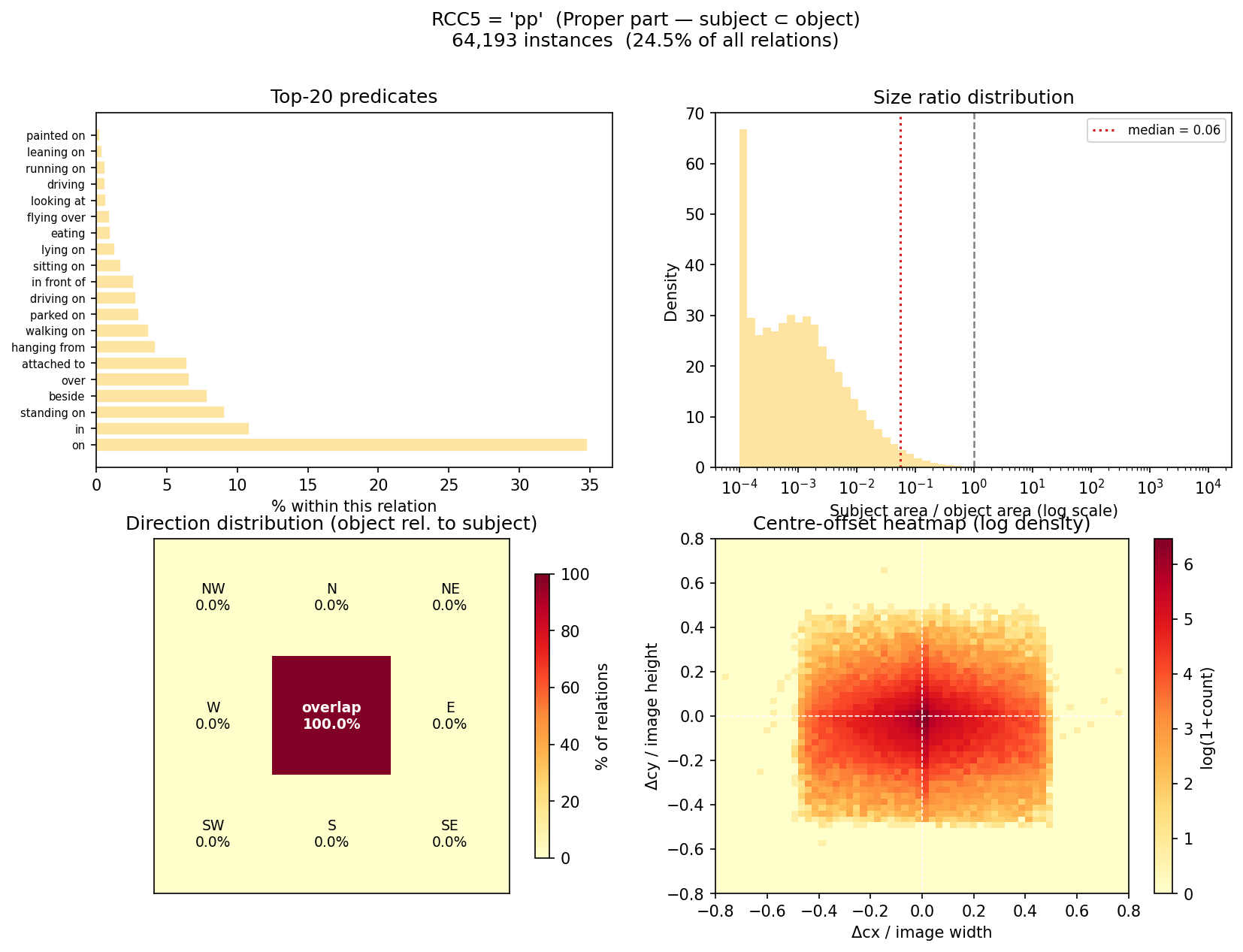}
  \caption{PSG predicates associated with the RCC5 \emph{proper part}
  (\texttt{pp}) topology -- the principal containment case.}
  \label{fig:rcc5-pp}
\end{figure}

Each ordered subject--object pair is assigned one of five jointly exhaustive,
pairwise disjoint topological relations of the RCC5 calculus
(Figure~\ref{fig:rcc5}), computed from the two bounding boxes
(Table~\ref{tab:rcc5}). The per-predicate empirical distributions
$\hat{P}(t\mid p)$ and their triplet variants are estimated by maximum
likelihood; a $(p,t)$ pair whose probability falls below the forbidden
threshold becomes a hard pruning constraint. The per-relation distributions
across PSG predicates are shown in
Figures~\ref{fig:rcc5-stacked}--\ref{fig:rcc5-pp}.

\begin{table}[ht]
  \centering
  \caption{\textbf{Region Connection Calculus (RCC)}. Topological Relations of the RCC5 fragment used to qualitatively model and apply geometric filtering.}
  \setlength{\tabcolsep}{8pt}
  \renewcommand{\arraystretch}{1.25}
  \sffamily
  \footnotesize
  \begin{tabular}{>{\columncolor{headerblue}}lll}
    \toprule\rowcolor{headerblue}
    \textbf{Atom} & \textbf{Relation} & \textbf{Description} \\
    \midrule
    \texttt{DR}  & Discrete            & Disjoint or touching; no shared interior \\
    \texttt{PO}  & Partial overlap     & Overlap, but neither contains the other \\
    \texttt{PP}  & Proper part         & Subject lies inside object \\
    \texttt{PPI} & Proper part inverse & Object lies inside subject \\
    \texttt{EQ}  & Equal               & The two regions coincide \\
    \bottomrule
  \end{tabular}
  \label{tab:rcc5}
\end{table}

\subsection{Cardinal direction}

\begin{figure}[t]
  \centering
  \includegraphics[width=0.5\linewidth]{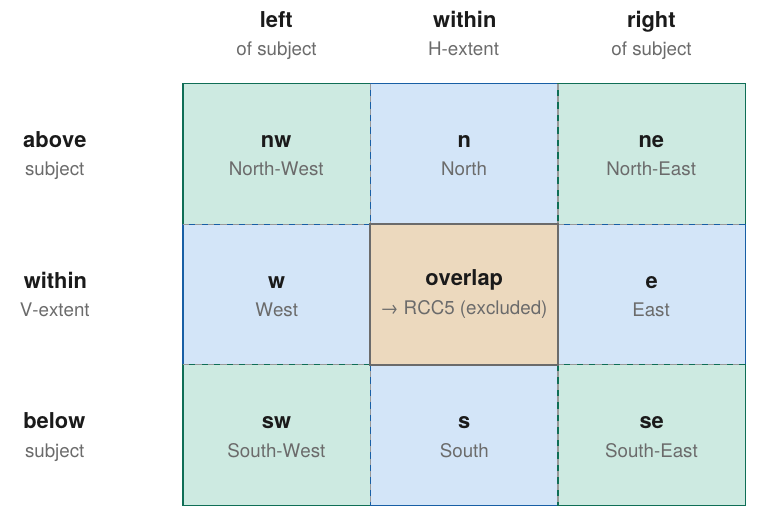}
  \caption{The eight cardinal directions used for geometric filtering. The
  central tile (object centroid within the subject's extents on both axes)
  is the topological-overlap case and is excluded from direction
  distributions.}
  \label{fig:direction}
\end{figure}

The plane around the subject bounding box is divided into nine tiles by
extending its edges; the object centroid selects one of eight cardinal
directions (Figure~\ref{fig:direction}, Table~\ref{tab:dir}). When the object
centroid falls within the subject's projected extent on both axes the pair is
topologically overlapping and is handled by the RCC5 relations above, so it is
excluded from the direction distributions. As with topology, per-predicate
and per-triplet distributions $\hat{P}(d\mid p)$ are estimated by maximum
likelihood and thresholded into forbidden direction constraints.

\begin{table}[ht]
  \centering
  \caption{The eight cardinal direction relations used for spatial filtering.
  Each relation is selected by the position of the object's centroid relative to
  the subject's projected extents on the horizontal and vertical axes.}
  \setlength{\tabcolsep}{8pt}
  \renewcommand{\arraystretch}{1.2}
  \sffamily
  \footnotesize
  \begin{tabular}{>{\columncolor{headerblue}}lll}
    \toprule\rowcolor{headerblue}
    \textbf{Atom} & \textbf{Direction} & \textbf{Condition (object centroid)} \\
    \midrule
    \texttt{n}  & North      & above subject, within horizontal extent \\
    \texttt{ne} & North-East & above and right of subject \\
    \texttt{e}  & East       & right of subject, within vertical extent \\
    \texttt{se} & South-East & below and right of subject \\
    \texttt{s}  & South      & below subject, within horizontal extent \\
    \texttt{sw} & South-West & below and left of subject \\
    \texttt{w}  & West       & left of subject, within vertical extent \\
    \texttt{nw} & North-West & above and left of subject \\
    \bottomrule
  \end{tabular}
  \label{tab:dir}
\end{table}

\subsection{Bounding-box features}

Beyond the topological and directional calculi, each pair is summarised by six
continuous geometric features, each partitioned into five bins
(Table~\ref{tab:bbox}). For every predicate (and triplet) the fraction of
training instances in each bin is recorded; bins whose fraction falls below
the forbidden threshold become finer-grained pruning constraints. The bin
name forms a single ASP atom.

\begin{table}[ht]
  \centering
  \caption{The six bounding-box features and their five-way binning. Offsets
  are normalised by image height/width; IoU and the containment fractions lie
  in $[0,1]$; the size ratio is subject area over object area.}
  \setlength{\tabcolsep}{2pt}
  \renewcommand{\arraystretch}{1.3}
  \sffamily
  \scriptsize
  \begin{tabular}{>{\columncolor{headerblue}}lll}
    \toprule\rowcolor{headerblue}
    \textbf{Feature} & \textbf{Quantity} & \textbf{Bins (low $\rightarrow$ high)} \\
    \midrule
    \texttt{dy}             & vertical centre offset   &
      \texttt{far\_above, above, aligned, below, far\_below} \\
    \texttt{dx}             & horizontal centre offset &
      \texttt{far\_left, left, aligned, right, far\_right} \\
    \texttt{iou}            & intersection-over-union  &
      \texttt{none, low, mid, high, very\_high} \\
    \texttt{size\_ratio}    & area ratio (subj/obj)    &
      \texttt{subj\_tiny, subj\_small, similar, subj\_large, subj\_huge} \\
    \texttt{contain\_s\_in\_o} & subject-in-object overlap &
      \texttt{none, low, mid, high, full} \\
    \texttt{contain\_o\_in\_s} & object-in-subject overlap &
      \texttt{none, low, mid, high, full} \\
    \bottomrule
  \end{tabular}
  \label{tab:bbox}
\end{table}

\noindent The bin cut-points are: \texttt{dy} and \texttt{dx} at
$\{-0.20, -0.05, 0.05, 0.20\}$; \texttt{iou} at $\{0.0, 0.05, 0.25, 0.50\}$;
\texttt{size\_ratio} at $\{0.25, 0.5, 2.0, 4.0\}$ (log-scaled around $1$); and
both containment fractions at $\{0.05, 0.30, 0.70, 0.95\}$.

\subsection{Functional rules}

A predicate is \emph{functional with degree $N$} if, in any image, at most $N$
distinct subjects relate to a given object through it. For each predicate the
empirical fan-in violation rate is measured: predicates with zero violations
at $N=1$ (\eg{} \emph{feeding}, \emph{guiding}, \emph{throwing} on PSG) are
tagged \emph{hard} and enforced as integrity constraints by the stage-2
program (Section~\ref{sec:functional}); predicates whose violation rate stays
below $5\%$ (confidence $\geq 0.95$) at a small fan-in budget are tagged
\emph{soft}. Functional rules are mined both per predicate and per triplet.
Table~\ref{tab:functional} lists a representative sample of soft functional
predicates.

\begin{table}[ht]
  \centering
  \caption{Representative soft functional rules on PSG. \emph{Fan-in} is the
  largest number of subjects observed for one object through the predicate;
  \emph{violation rate} is the fraction of object occurrences exceeding it.}
  \setlength{\tabcolsep}{8pt}
  \renewcommand{\arraystretch}{1.2}
  \sffamily
  \footnotesize
  \begin{tabular}{>{\columncolor{headerblue}}lccc}
    \toprule\rowcolor{headerblue}
    \textbf{Predicate} & \textbf{Instances} & \textbf{Fan-in} &
      \textbf{Violation rate} \\
    \midrule
    carrying  & $2{,}297$  & $2$ & $0.0004$ \\
    swinging  & $721$      & $2$ & $0.0014$ \\
    wearing   & $2{,}834$  & $2$ & $0.0035$ \\
    driving   & $584$      & $3$ & $0.0051$ \\
    holding   & $10{,}037$ & $4$ & $0.0065$ \\
    riding    & $1{,}913$  & $4$ & $0.0272$ \\
    \bottomrule
  \end{tabular}
  \label{tab:functional}
\end{table}

\subsection{Relational properties}
\label{sec:relprop}

Three syntactic relational-property patterns are mined at the predicate level:
\emph{symmetric} $p(x,y)\Rightarrow p(y,x)$; \emph{inverse} $p(x,y)\Rightarrow
q(y,x)$ for $q\neq p$; and \emph{composition} $p_1(x,z)\wedge p_2(z,y)
\Rightarrow p_h(x,y)$. Each candidate is stored with its support and
confidence (mined with a minimum confidence of $0.8$ for symmetric/inverse
and $0.5$ for composition). These patterns capture logical regularities that
hold independently of any single image; the composition rules are compiled
into \texttt{horn\_composition/3} background facts and used in stage~1 to
bridge two network candidates. The atom name \texttt{horn\_composition} is retained for backward compatibility
with the codebase.

\section{Filter Behaviour on PSG}
\label{app:behaviour}

This section reports an empirical analysis of how the filter behaves on the
PSG test split: which rules fire (Section~\ref{sec:firing}), what effect they
have on ground-truth predictions (Section~\ref{sec:gt-impact}), and how each
rule type contributes to the aggregate F1@K score
(Section~\ref{sec:ablation}).

\subsection{Which rules fire?}
\label{sec:firing}

\begin{figure}[t]
  \centering
  \begin{minipage}[t]{0.49\linewidth}
    \centering
    \includegraphics[width=\linewidth]{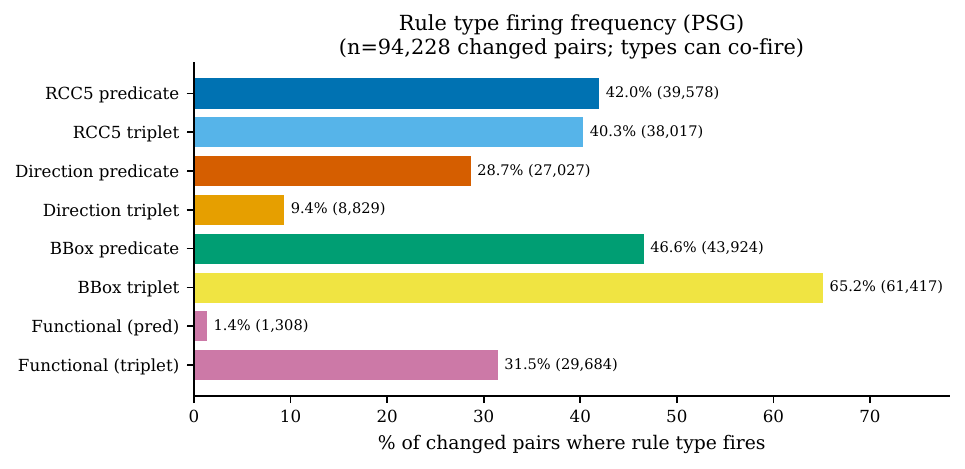}
  \end{minipage}\hfill
  \begin{minipage}[t]{0.49\linewidth}
    \centering
    \includegraphics[width=\linewidth]{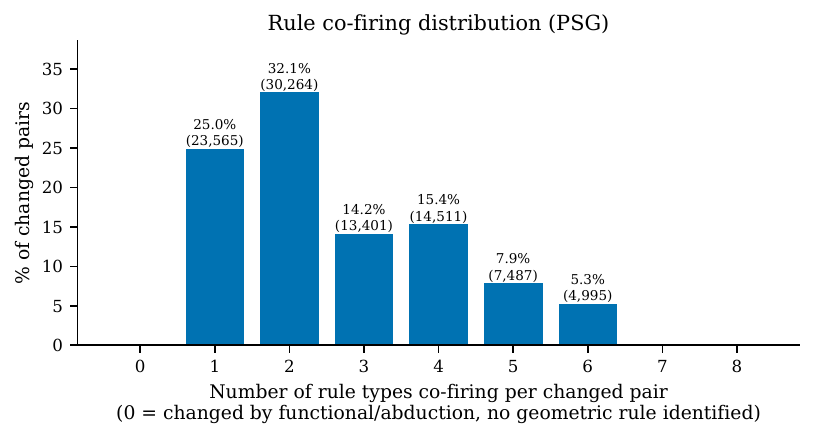}
  \end{minipage}
  \caption{Rule firing on PSG. \emph{Left:} firing frequency of each rule
  type, expressed as the percentage of changed pairs in which a rule of that
  type fires (types can co-fire, so percentages sum to more than $100$).
  \emph{Right:} distribution of the number of rule types co-firing per
  changed pair; roughly half of all changed pairs trigger three or more rule
  types simultaneously.}
  \label{fig:firing-supp}
\end{figure}

Figure~\ref{fig:firing-supp}~(left) summarises rule-type firing frequency. Across
the $94\,228$ pairs whose predicate changes under the filter, bounding-box
features (predicate and triplet level combined) fire on roughly two thirds of
changed pairs, RCC5 on about $40\%$, direction on about $30\%$, and
functional on a third. Pred-level and triplet-level rules can co-fire on the
same pair, so the percentages sum to more than $100$. The co-firing
distribution (Figure~\ref{fig:firing-supp}, right) confirms that this overlap is
the rule rather than the exception: $32.1\%$ of changed pairs trigger exactly
two rule types and $42.8\%$ trigger three or more, with only a quarter of
changes driven by a single rule type.

\begin{figure}[t]
  \centering
  \includegraphics[width=0.85\linewidth]{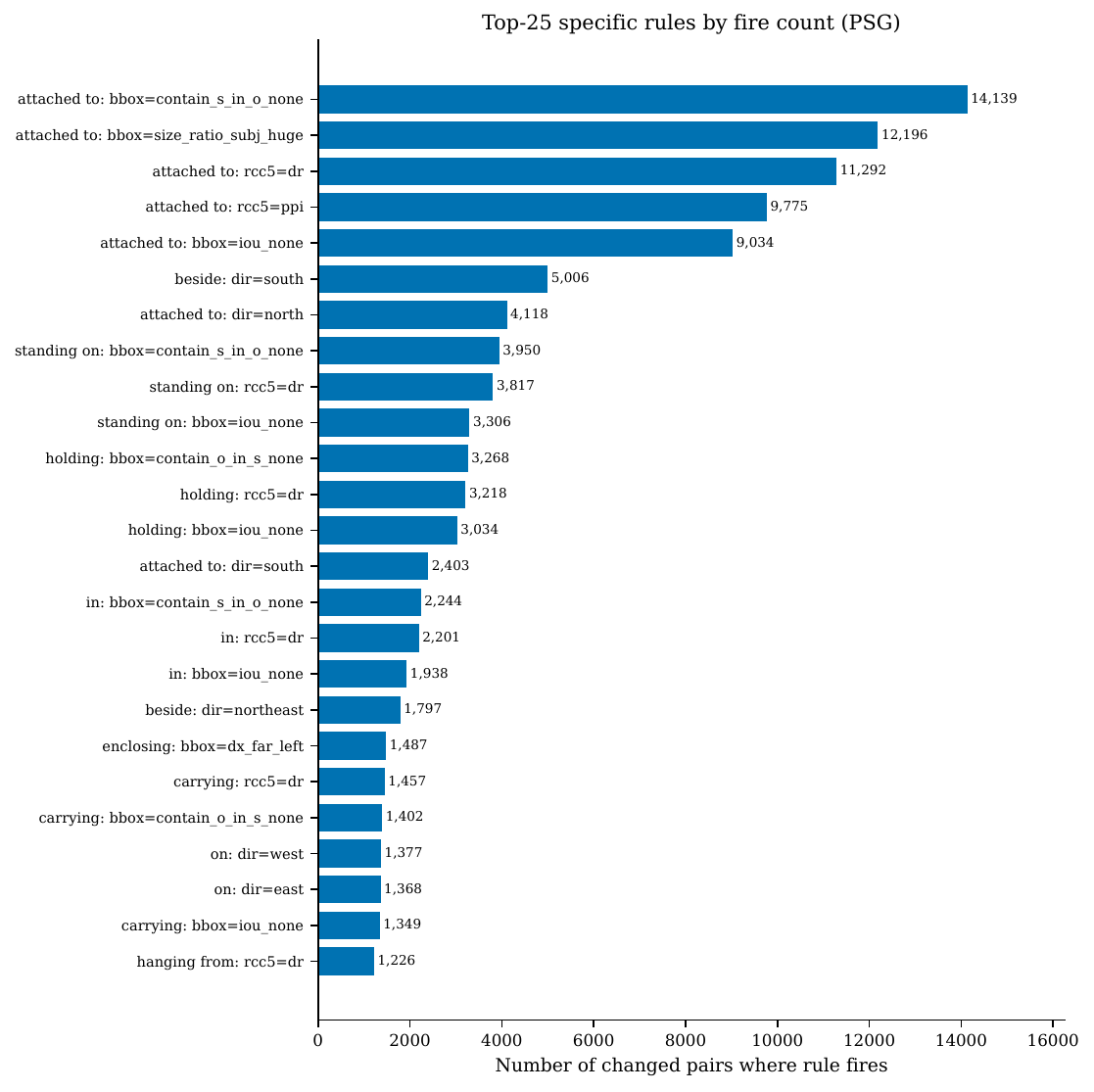}
  \caption{The 25 most-frequently firing specific rules on PSG. Rules
  involving \emph{attached to} dominate, reflecting the predicate's strong
  bounding-box signature (one object contained or aligned within the other).}
  \label{fig:top-rules}
\end{figure}

Figure~\ref{fig:top-rules} drills down to individual rules: the 25
single-condition rules that fire most often. Rules involving \emph{attached
to} dominate, reflecting the strong bounding-box signature of that predicate
(typically containment or near alignment). Several predicates appear more
than once with complementary conditions (\eg{} \emph{attached to} with
\texttt{bbox=contain\_s\_in\_o=none}, \texttt{bbox=size\_ratio=subj\_huge},
\texttt{rcc5=dr}, and \texttt{rcc5=ppi}), showing that the filter does not
rely on a single trigger but on a small set of mutually reinforcing
constraints.

\subsection{Effect on ground-truth predictions}
\label{sec:gt-impact}

\begin{figure}[t]
  \centering
  \includegraphics[width=0.95\linewidth]{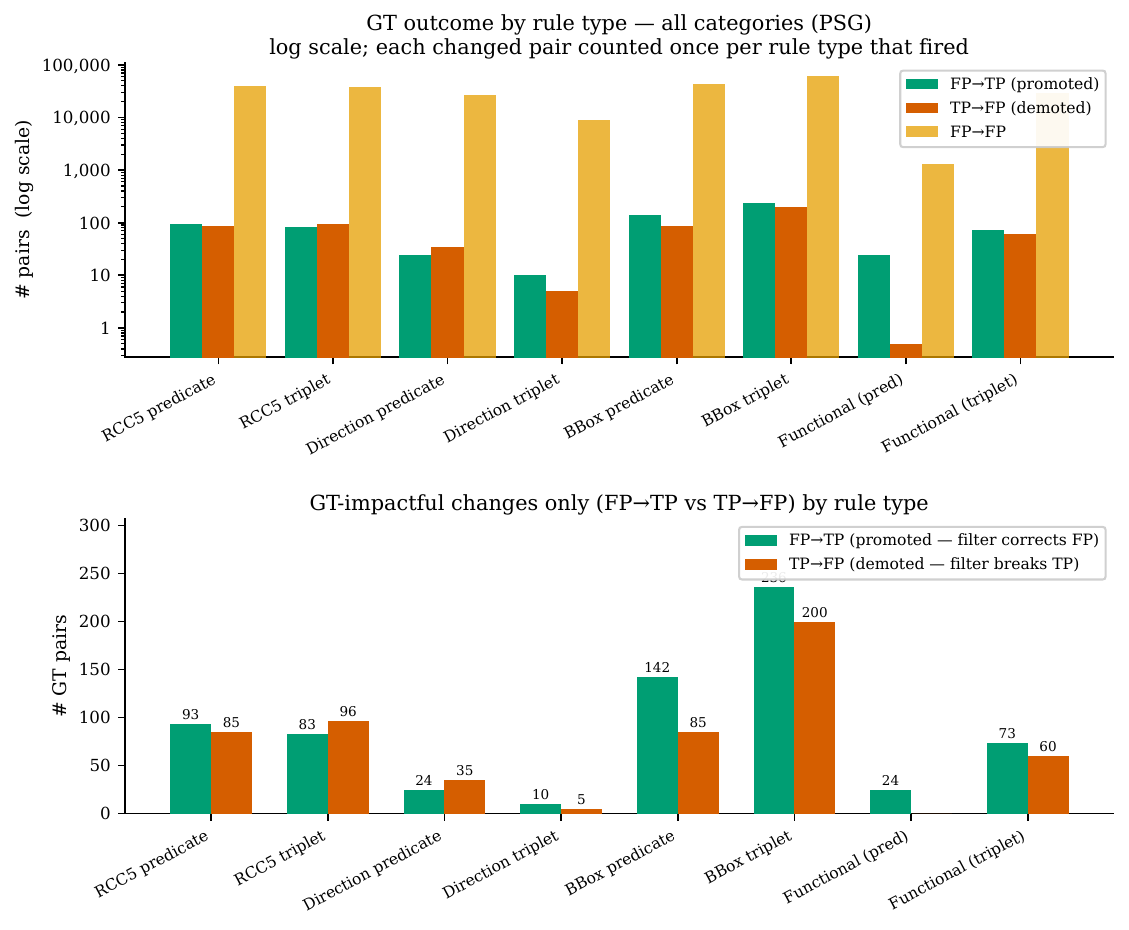}
  \caption{Ground-truth outcome by rule type on PSG. Top: full distribution
  on a log scale; the vast majority of changes are FP$\to$FP (neutral
  rearrangements between two wrong predicates). Bottom: GT-impactful changes
  only. Bounding-box triplet rules have the largest positive contribution
  (FP$\to$TP) but also the largest collateral damage (TP$\to$FP); functional
  rules at predicate level are almost purely beneficial.}
  \label{fig:gt-outcome}
\end{figure}

Whether a rule fires often is not the same as whether it helps. We measure
each rule type's \emph{GT outcome} by labelling every changed pair as one of:
\textbf{FP$\to$TP} (the filter corrects a false-positive network prediction
into the GT predicate), \textbf{TP$\to$FP} (the filter demotes a correct
prediction), or \textbf{FP$\to$FP} (both predictions were already wrong;
neutral). Figure~\ref{fig:gt-outcome} reports both the full distribution and
the GT-impactful subset.

\begin{figure}[t]
  \centering
  \includegraphics[width=0.95\linewidth]{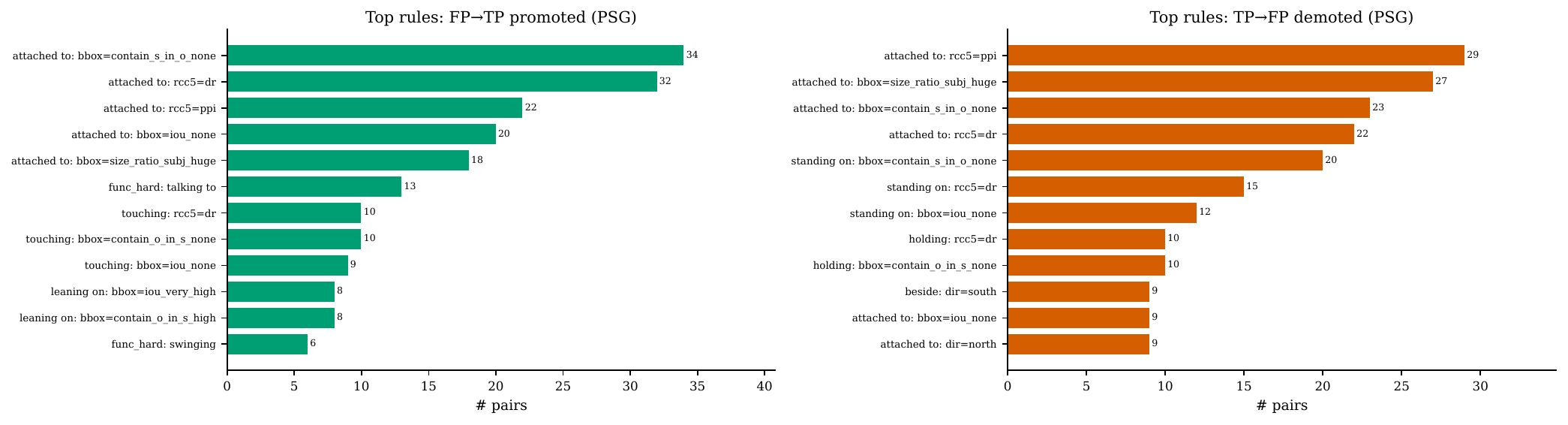}
  \caption{The single rules most responsible for GT-impactful changes on
  PSG. Left: rules that most often promote a false-positive prediction to
  the GT predicate. Right: rules that most often demote a correct
  prediction.}
  \label{fig:top-rules-by-gt}
\end{figure}

Figure~\ref{fig:top-rules-by-gt} identifies the specific rules behind these
gains and losses. The same \emph{attached to} rule family dominates both
columns: highly active rules whose net effect is determined by triplet-level
context. This is the empirical justification for keeping triplet-level
constraints alongside predicate-level ones rather than relying on either in
isolation.

\subsection{Aggregate F1 contribution}
\label{sec:ablation}

\begin{figure}[t]
  \centering
  \includegraphics[width=0.95\linewidth]{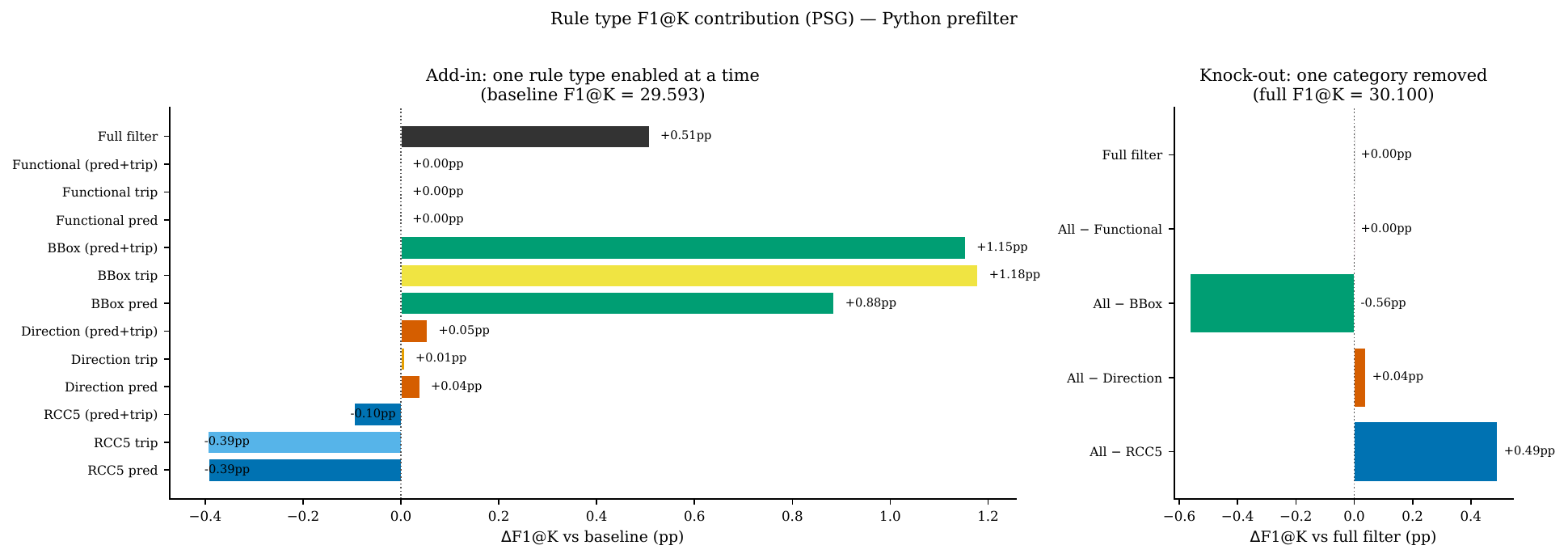}
  \caption{F1@K contribution per rule type on PSG.}
  \label{fig:f1-contrib}
\end{figure}

Figure~\ref{fig:f1-contrib} reports an \emph{add-in / knock-out} ablation:
enabling one rule type at a time on top of the unfiltered baseline (left),
and removing one rule type from the full filter (right). The bounding-box
family (predicate plus triplet) is the dominant positive contributor at
$+1.18$pp; direction rules add little on their own. RCC5 rules in isolation
slightly hurt the score because they over-prune, but the full filter recovers
this loss thanks to the network-first selection and the spatial-support
fallback. The knock-out panel confirms the picture: removing the bounding-box
family costs $-0.56$pp; removing RCC5 actually improves the score by
$+0.49$pp, suggesting that the RCC5 family is partially redundant with the
finer-grained bounding-box features in the full configuration.

\section{Extended Figures}\label{app:figures}

This section collects supporting figures for the PSG dataset: dataset-level
geometric statistics, per-RCC5-topology analyses for the two most
discriminative topologies, and additional per-predicate diagnostics.

\begin{figure}[htbp]
  \centering
  \includegraphics[width=0.98\linewidth]{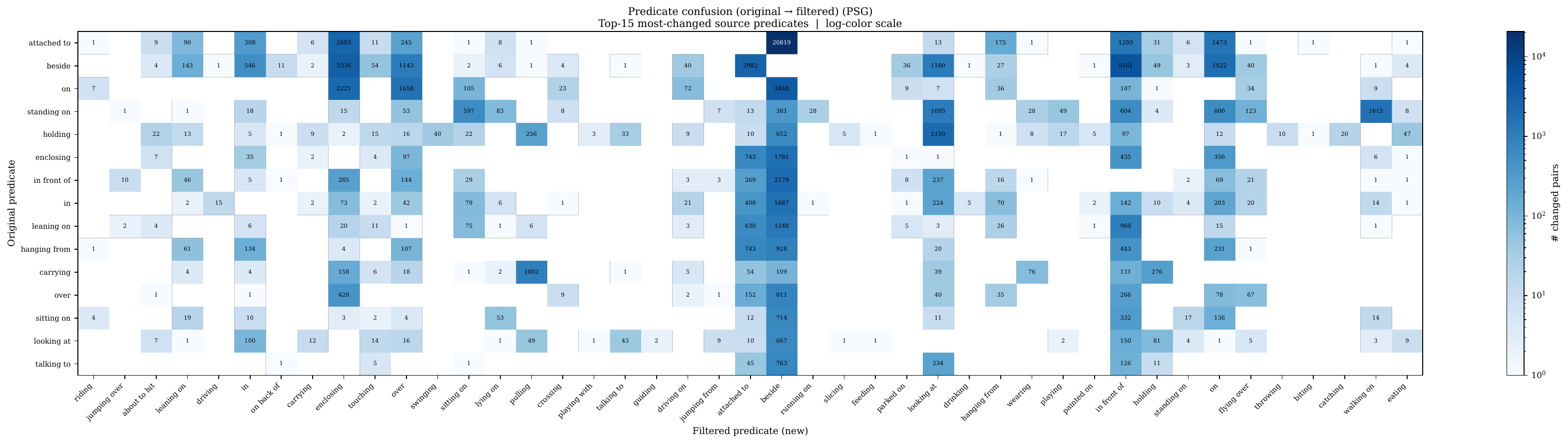}
  \caption{Predicate confusion matrix (original $\to$ filtered) for the top-15
  most-changed source predicates on PSG. Cell values are the number of pairs
  whose predicate was reassigned from row to column; the colour scale is
  logarithmic.}
  \label{fig:pred-confusion}
\end{figure}

\begin{figure}[htbp]
  \centering
  \includegraphics[width=\linewidth]{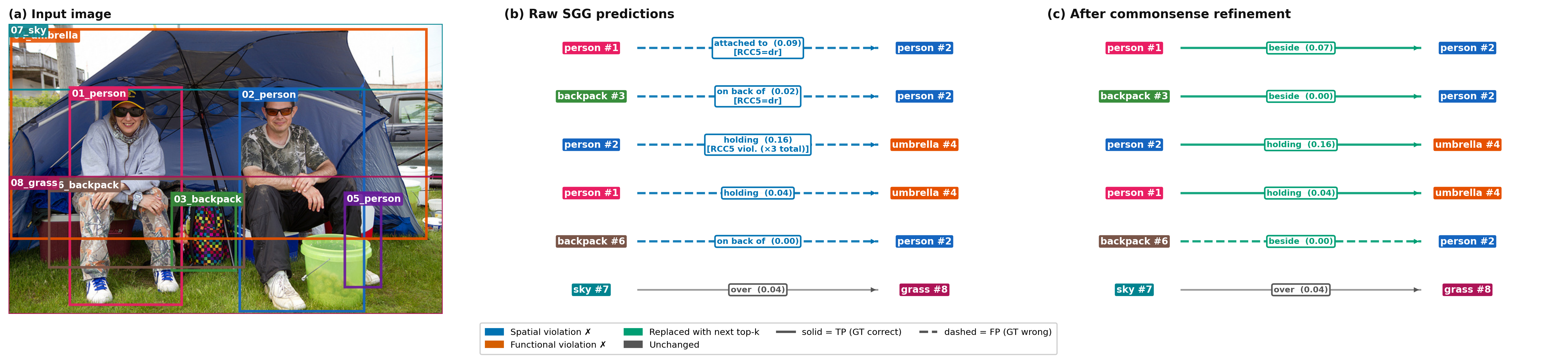}\\[6pt]
  \includegraphics[width=\linewidth]{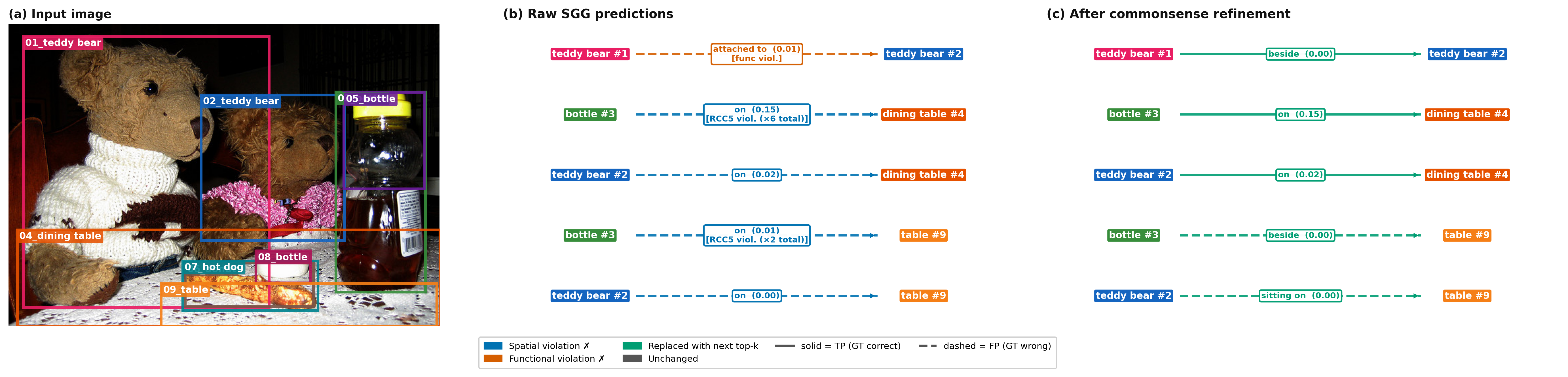}\\[6pt]
  \includegraphics[width=\linewidth]{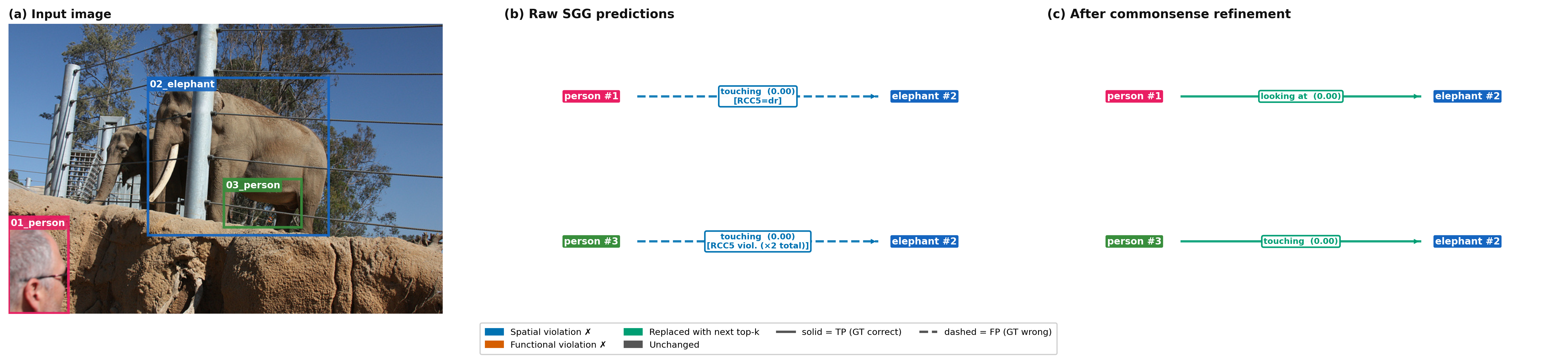}\\[6pt]
  \includegraphics[width=\linewidth]{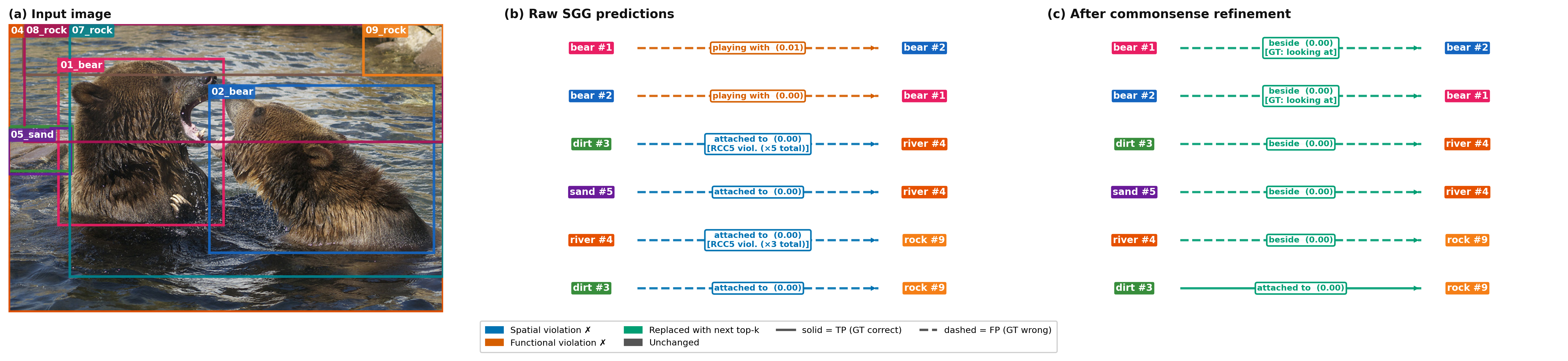}\\[6pt]
  \includegraphics[width=\linewidth]{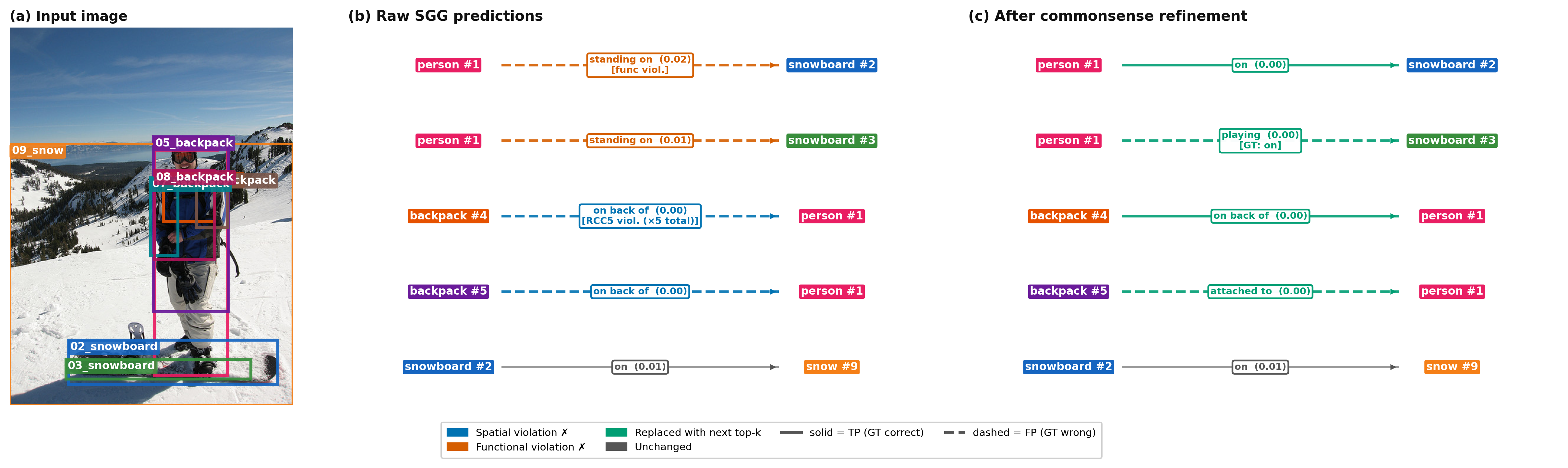}

  \caption{Qualitative examples of commonsense refinement on PSG. For each
  image: \emph{(a)} input, \emph{(b)} raw SGG predictions, \emph{(c)} after
  refinement. Dashed edges are false positives, solid edges true positives;
  blue/orange tags mark spatial/functional violations, green marks
  next-top-$k$ replacements.}
  \label{fig:qual-1}
\end{figure}

\begin{figure}[htbp]
  \centering
  \includegraphics[width=0.95\linewidth]{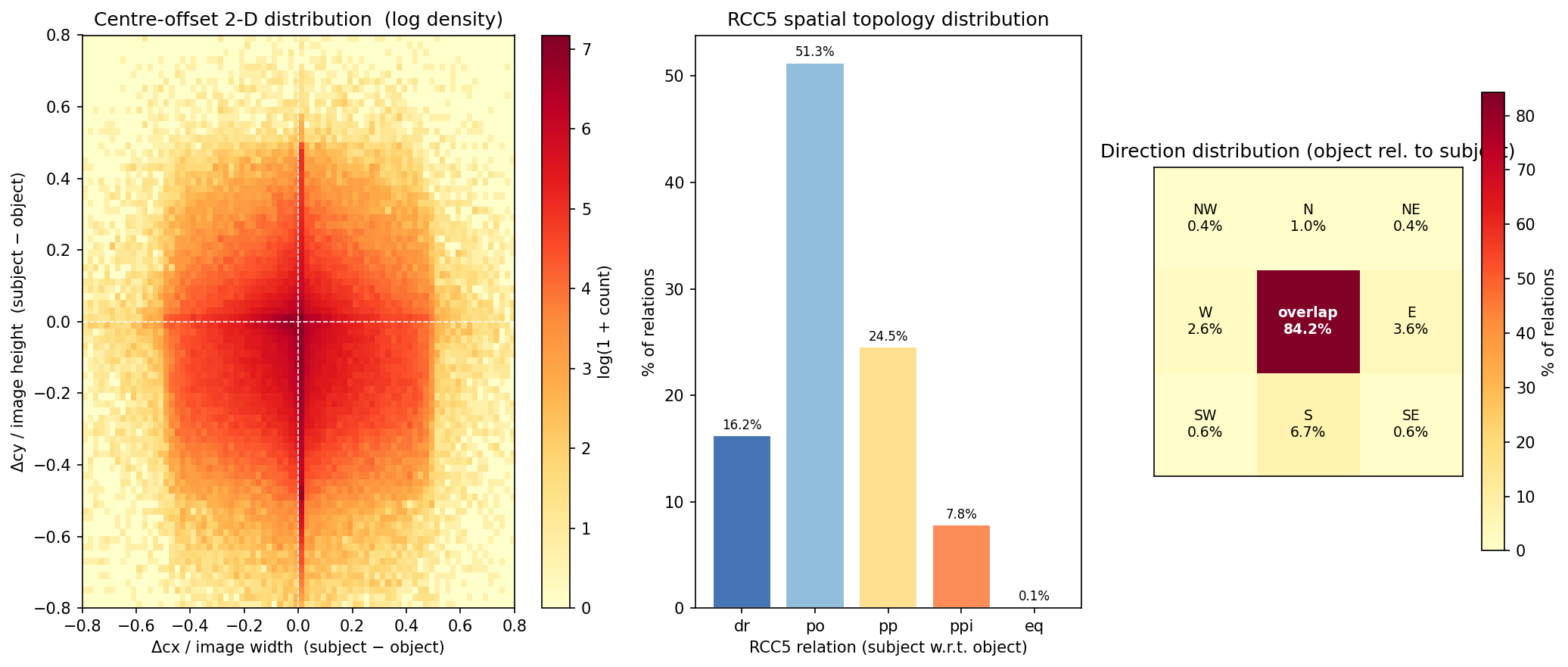}
  \caption{Geometric statistics of subject--object pairs for the PSG dataset.}
  \label{fig:geometric-stats}
\end{figure}

\begin{figure}[htbp]
  \centering
  \includegraphics[width=0.95\linewidth]{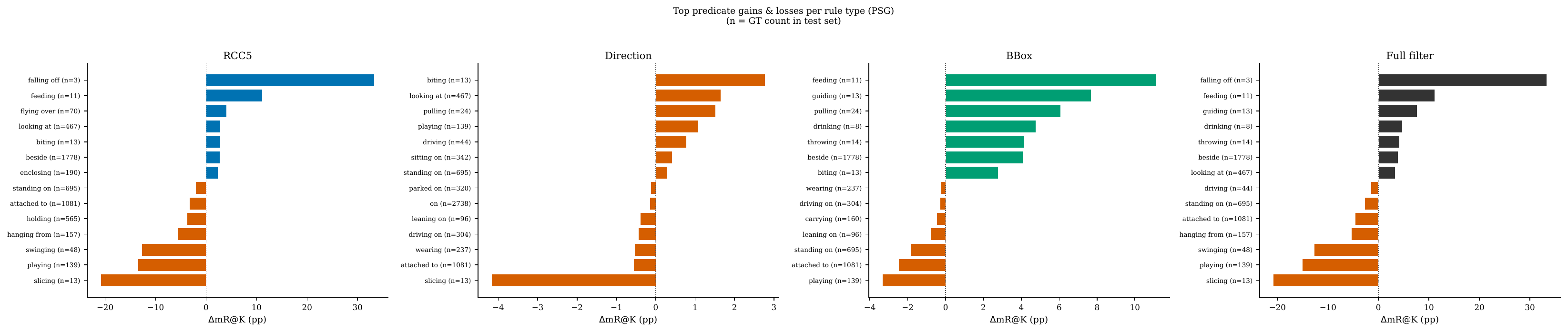}
  \caption{Per-predicate gains and losses by rule type on PSG, measured as
  the change in mR@K (percentage points). Each panel restricts attention to
  one rule type; \emph{Full filter} (right) is the combined effect.}
  \label{fig:top-pred-changes}
\end{figure}

\begin{figure}[htbp]
  \centering
  \includegraphics[width=0.85\linewidth]{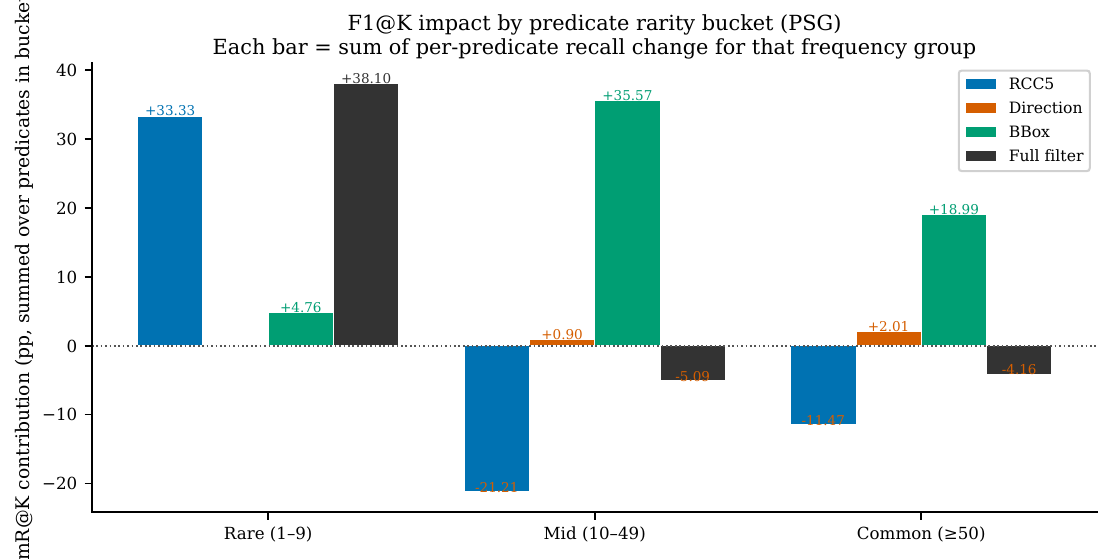}
  \caption{F1@K contribution per rule type on PSG, grouped by predicate
  rarity bucket (Rare: $1$--$9$ GT instances; Mid: $10$--$49$; Common:
  $\geq 50$).}
  \label{fig:bucket-f1}
\end{figure}

\begin{figure}[htbp]
  \centering
  \includegraphics[width=0.95\linewidth]{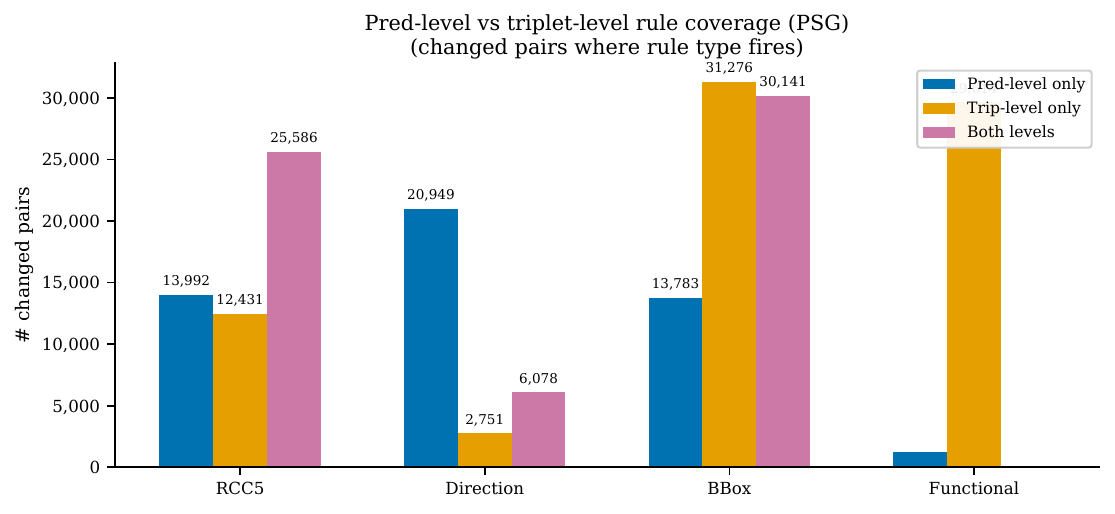}
  \caption{Coverage of predicate-level versus triplet-level rules per rule
  type on PSG, in terms of how many changed pairs are touched by rules at
  each granularity.}
  \label{fig:pred-trip-breakdown}
\end{figure}

\end{document}